\documentclass[twocolumn]{article}
\usepackage{arxiv}
\usepackage{cite}
\usepackage[utf8]{inputenc}

\usepackage[colorlinks = true,
linkcolor = blue,
citecolor = blue,
anchorcolor = blue,
urlcolor = blue]{hyperref}      
\usepackage{booktabs}       
\usepackage{amsfonts}       
\usepackage{nicefrac}       
\usepackage{microtype}      
\usepackage{amsmath}  
\usepackage{graphicx}
\usepackage{float}
\usepackage{enumerate}
\usepackage{enumitem}
\usepackage{array}  
\usepackage[font=small]{caption}
\usepackage{subcaption}
\date{}
\renewcommand{\headeright}{}
\renewcommand{\undertitle}{}

\title{Advancing Machine Learning in Industry 4.0: Benchmark Framework for Rare-event Prediction in Chemical Processes}

\author{
    \href{https://orcid.org/0000-0003-2361-4032}{\includegraphics[scale=0.06]{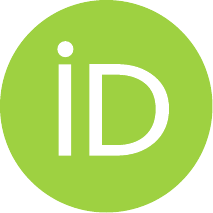}\hspace{1mm}Vikram Sudarshan} \\
    Department of Chemical and Biomolecular Engineering\\
    University of Pennsylvania\\
    Philadelphia, PA 19104-6393 \\
    \texttt{vsud@seas.upenn.edu} \\
    \And
    \href{https://orcid.org/0000-0002-4217-8670}{\includegraphics[scale=0.06]{orcid.pdf}\hspace{1mm}Warren D.~Seider}\thanks{} \\
    Department of Chemical and Biomolecular Engineering\\
    University of Pennsylvania\\
    Philadelphia, PA 19104-6393 \\
    \texttt{seider@seas.upenn.edu} \\
}

\begin{document}

\twocolumn[
\maketitle
\begin{abstract}
\vspace{1\baselineskip}  
Previously, using forward-flux sampling (FFS) and machine learning (ML), we developed multivariate alarm systems to counter rare un-postulated abnormal events. Our alarm systems utilized ML-based predictive models to quantify committer probabilities as functions of key process variables (e.g., temperature, concentrations, and the like), with these data obtained in FFS simulations. Herein, we introduce a novel and comprehensive benchmark framework for rare-event prediction, comparing ML algorithms of varying complexity, including Linear Support-Vector Regressor and k-Nearest Neighbors, to more sophisticated algorithms, such as Random Forests, XGBoost, LightGBM, CatBoost, Dense Neural Networks, and TabNet. This evaluation uses comprehensive performance metrics, such as: \textit{RMSE}, model training, testing, hyperparameter tuning and deployment times, and number and efficiency of alarms. These balance model accuracy, computational efficiency, and alarm-system efficiency, identifying optimal ML strategies for predicting abnormal rare events, enabling operators to obtain safer and more reliable plant operations.
\end{abstract}

\vspace{1\baselineskip}  
\keywords{Benchmark Framework, Machine Learning, Rare Abnormal Events, Alarm-system Efficiency}
\vspace{2\baselineskip}
]
{
  \renewcommand{\thefootnote}%
    {\fnsymbol{footnote}}
  \footnotetext[1]{Corresponding author
}
}
\section{Introduction}
\subsection{Progress of Industrial Revolution and Challenge of Rare Un-postulated Abnormal Events}
\label{sec_1.1}
Over the past few centuries, several industrial revolutions have transformed the chemical and manufacturing industries. These began with the first industrial revolution in the mid-1700s, focused on mechanization through water and steam engines and railroads \cite{crafts_explaining_2011, mohajan_first_2019}; then moved to the second industrial revolution in the mid-to-late 1800s, focused on electrification, ramping-up manufacturing, and improving efficiency by introducing assembly lines \cite{mokyr_second_1998}; then advanced to the third industrial revolution in the late-1900s, introducing automation technologies (e.g., distributed control systems; i.e., DCS), computers, and electronics \cite{mohajan_third_2021, naboni_third_2015}; and then proceeded to the current Industry 4.0 vision of digitalization consisting of path-breaking technologies such as the internet-of-things (IoT) \cite{belli_toward_2019, soori_internet_2023}, artificial intelligence and machine learning (AI/ML) \cite{becue_artificial_2021, candanedo_machine_2018, dingli_artificial_2021}, cybersecurity and cyber-physical systems \cite{culot_addressing_2019, ervural_overview_2018, hashimoto_safety_2013}, and big-data analytics and cloud computing \cite{gokalp_big_2016, kim_review_2017}.  Numerous perspectives are anticipating Industry 5.0, with foci on customization and sustainability, consisting of technologies such as human-computer interaction, collaborative robotics, and, augmented reality and mixed reality (AR/MR) \cite{barata_industry_2023, demir_industry_2019, ghobakhloo_behind_2023,raja_santhi_industry_2023}.

\paragraph{}Remarkably, despite these breakthroughs, the chemical manufacturing industries struggle to prevent safety accidents (e.g., thermal runaways, release of flammables, and chemical spillage) and reliability failure events (e.g., poor product quality and related financial losses).  The former have resulted in numerous fatalities, including: the Pemberton Mill accident in 1860, the Grover shoe factory disaster in 1905, the Flixborough disaster in 1974 \cite{hailwood_learning_2016}, the Bhopal gas tragedy in 1984 \cite{broughton_bhopal_2005, gupta_bhopal_2002, sriramachari_bhopal_2004}, the Chernobyl disaster in 1986 \cite{saenko_chernobyl_2011}, the BP Texas City refinery explosion in 2005 \cite{holmstrom_csb_2006}, the Deepwater Horizon oil spill in 2010 \cite{beyer_environmental_2016}, and the Fukushima disaster in 2011 \cite{labib_learning_2015}.  Often, such catastrophic accidents are triggered by rare, \textit{un-postulated} abnormal events or process faults unidentified at the time of occurrence.  Additionally, there are very few occurrence data, making it challenging to predict their likelihood using data-driven quantitative techniques.  While extensive near-miss data often help to prevent accidents, more accurate estimates are needed.  Moreover, routine alarm management systems, created using HAZOP studies,  are often unable to identify such abnormal rare events; e.g., the root-cause of the BP Texas City refinery explosion was not identified during HAZOP studies \cite{us_chemical_safety_and_hazard_investigation_board_bp_2007}.  While automated Safety Instrumented Systems (SIS) are usually successful in preventing accidents through interlock activation, they contribute to plant reliability issues (i.e., causing shutdowns, maintenance, and start-up), resulting in production-time and financial losses.  \textbf{Given these numerous challenges, there is a strong motivation to develop enhanced multivariate alarm systems for identifying and handling these rare un-postulated abnormal events more-efficiently – enabling operators to improve plant safety and reliability.} 

\subsection{Artificial Intelligence and Machine Learning (AI/ML) for Quantitative Analyses of Rare Events
}
\label{sec_1.2}
AI/ML is one of the cornerstones of Industry 4.0's vision for improved automation through digital transformation.  Over the past decade, there has been an exponential rise in AI/ML research across several scientific domains, including chemical engineering applications: drug discovery \cite{lavecchia_machine-learning_2015, vamathevan_applications_2019}, catalysis \cite{kitchin_machine_2018, toyao_machine_2020}, materials science \cite{morgan_opportunities_2020, wei_machine_2019}, computational fluid dynamics \cite{hanna_machine-learning_2020,kochkov_machine_2021}, molecular dynamics \cite{gastegger_machine_2017, wang_machine_2020}, process monitoring and fault detection \cite{arunthavanathan_machine_2022, harkat_machine_2020}, to name a few.  With respect to quantitative estimation of rare-events for chemical process safety, AI/ML-based techniques have been developed; a parametric reduced-order modeling approach was developed to estimate and analyze the consequence of rare abnormal events, using the k-Nearest Neighbors ML algorithm, and demonstrated on a cardon dioxide release study \cite{kumari_development_2021}.  Additionally, optimal ML algorithms were applied to predict and analyze the root-causes of occupational safety events \cite{sarkar_application_2019}.  In related work, three categories of classification ML algorithms; i.e., wide, deep, and wide and deep, were introduced and analyzed using accident data for severity predictions \cite{tamascelli_learning_2022}.  Moreover, a novel anomaly detection-based classification algorithm was developed using real-time data from industrial processes \cite{quatrini_machine_2020}. 
\paragraph{}Despite significant advances, the utilization of ML algorithms for prediction of rare abnormal events presents significant concerns.  From amongst a vast choice of ML techniques, it is crucial to select an algorithm most relevant to the target application.  Additionally, most ML algorithms developed for rare events are purely data-driven; i.e., based on data from process historians, accident data, or alarm databases. And, due to the scarcity of data for truly rare events, data quality is a concern, given that ML model performance relies heavily on such data \cite{budach_effects_2022, jain_overview_2020}.  \textbf{Given this lack of occurrence data, it is important to integrate AI/ML-based techniques with efficient simulation-based techniques (e.g., path-sampling), capable of identifying and generating pathways for rare un-postulated abnormal events.}

\subsection{Benchmark Analyses of ML Algorithms
}
\label{sec_1.3}
With the challenge in selecting relevant algorithms, benchmark analyses of ML algorithms are ubiquitous across several scientific domains having access to open-source databases.  A large-scale benchmark framework; i.e., MoleculeNet, was developed for benchmarking ML algorithms for molecular datasets, including data for over 700,000 compounds \cite{wu_moleculenet_2018}.  Similar analyses and comparisons among several ML algorithms have been conducted for traffic-sign recognition \cite{stallkamp_man_2012}, healthcare datasets \cite{purushotham_benchmarking_2018}, federated learning \cite{he_fedml_2020}, scientific machine learning \cite{thiyagalingam_scientific_2022}, detection of software defects \cite{aleem_benchmarking_2015}, time-series forecasting \cite{pfisterer_benchmarking_2021, xie_benchmarking_2020}, and cancer research \cite{feltes_cumida_2019}, to name a few.  Such rigorous benchmark analyses have also been extended specifically for tabular data; i.e., the most common data format utilized across several scientific domains \cite{shwartz-ziv_tabular_2022}.  Many of these studies report that for supervised learning tasks (i.e., regression and classification) using tabular data, gradient-boosting frameworks (e.g., XGBoost, CatBoost, LightGBM) outperform more-complex neural network-based, deep-learning architectures, achieving comparable or superior accuracies at lower computational costs \cite{borisov_deep_2022, grinsztajn_why_2022, shwartz-ziv_tabular_2022, uddin_confirming_2024}.  More-recently, in related research, a comprehensive survey was conducted for predicting rare-events – considering data, preprocessing, algorithmic techniques, and evaluations \cite{shyalika_comprehensive_2023}.
\paragraph{}While most studies include several datasets and algorithms, it is very challenging to extend these for data concerning safety and reliability of chemical processes – due to lack of occurrence data accompanying such rare-events.  \textbf{Additionally, apart from model accuracies/errors and computational costs, it is also crucial to analyze the impact of ML algorithms on alarm-system efficiency; e.g., the number and efficiency of alarms annunciated in identifying abnormal behavior accurately – a missing component in existing benchmark studies.}

\subsection{Prior Research: Developing Improved, Multivariate Alarm Systems Using Forward-Flux Sampling and Machine Learning
}
\label{sec_1.4}
Given the limitations of HAZOP-based alarm management systems in identifying and mitigating rare un-postulated events, in previous research, we developed improved, novel, multivariate alarm systems using forward-flux sampling (FFS) and machine learning – based on random statistical “noise”-induced perturbations in one or more process variables that ultimately result in rare un-postulated abnormal shifts from normal to undesirable (i.e., \textit{unsafe} or \textit{unreliable}) regions \cite{sudarshan_understanding_2021, sudarshan_multivariate_2023}.  Our alarm systems utilize ML-based algorithms that predict the \textit{committer probability} as a function of the process variables.  Then, to enhance the quality and efficiency of the alarm systems, we developed an integrated framework for alarm rationalization and dynamic risk analyses \cite{sudarshan_alarm_2024}.  First, our techniques were demonstrated successfully for a relatively simple exothermic CSTR process model.  Then, we improved our methods for more-complex polymerization CSTRs, resulting in \textit{dynamic}, \textit{bidirectional} multivariate alarm systems based on real-time predictions of committer probabilities, using more-advanced nonparametric ML algorithms.  This addressed the \textit{decision-science} component of risk assessment and machine learning; i.e., given predictions by the ML algorithms, determining the actionable strategies for reducing the real-time committer probabilities \cite{sudarshan_path-sampling_2024}. 

\begin{figure*}[t]
    \centering
    \includegraphics[width=0.65\linewidth]{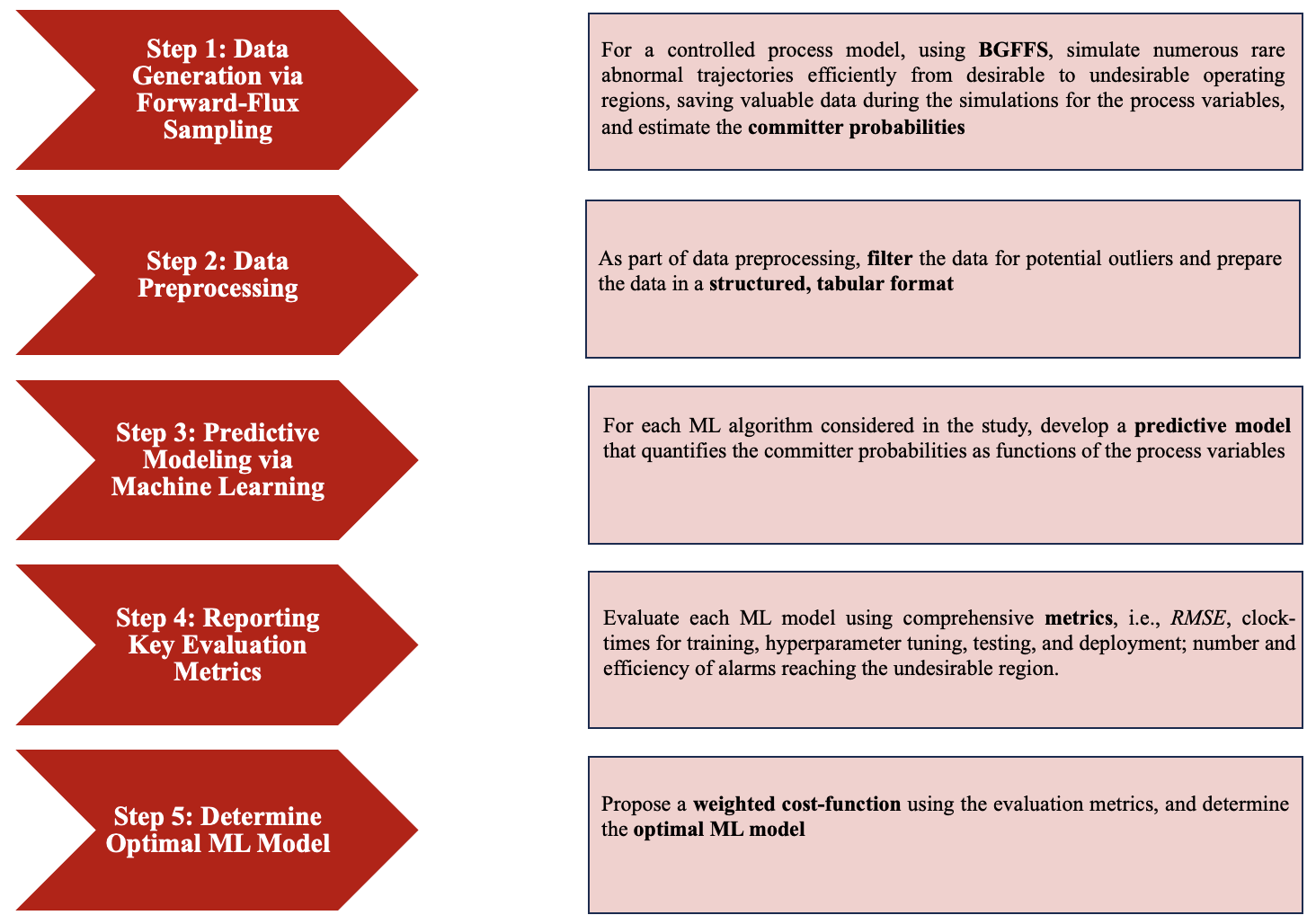}
    \caption{Overview of Key Steps}
    \label{fig_1}
\end{figure*}

\subsection{Benchmark Analyses of ML Algorithms for Rare Abnormal Events
}
\label{sec_1.5}
In this paper, we introduce a comprehensive framework for benchmark analyses, comparing several ML algorithms, of varying complexities, for un-postulated rare-event predictions of chemical process models.  We begin with Linear Support-Vector Regressor (Linear SVR), k-Nearest Neighbors (kNNs), and move to more-complex algorithms; i.e., gradient-boosted decision trees (i.e., XGBoost, LightGBM, CatBoost) and deep-learning approaches (i.e., dense neural networks and TabNet).  Two chemical process models are considered; i.e., a PI-controlled exothermic CSTR, and a PID-controlled polystyrene CSTR, using five tabular datasets for committer probability-process variables data generated using the branched-growth variant of FFS (i.e., BG-FFS).  In our evaluation, several metrics are considered, including: \textit{RMSE}, clock-times recorded for training, testing, hyperparameter tuning and model deployment, and factors affecting alarm systems, including \textit{number} and \textit{efficiency} of multivariate alarms activated in real-time based on the predictions provided by each ML algorithm.  By considering diverse evaluation metrics (i.e., with alarm efficiency being novel when benchmarking ML algorithms), we seek to identify optimal ML strategies to predict and handle rare un-postulated abnormal events, thereby, improving overall safety and reliability.

\begin{figure}
    \centering
    \includegraphics[width=1\linewidth]{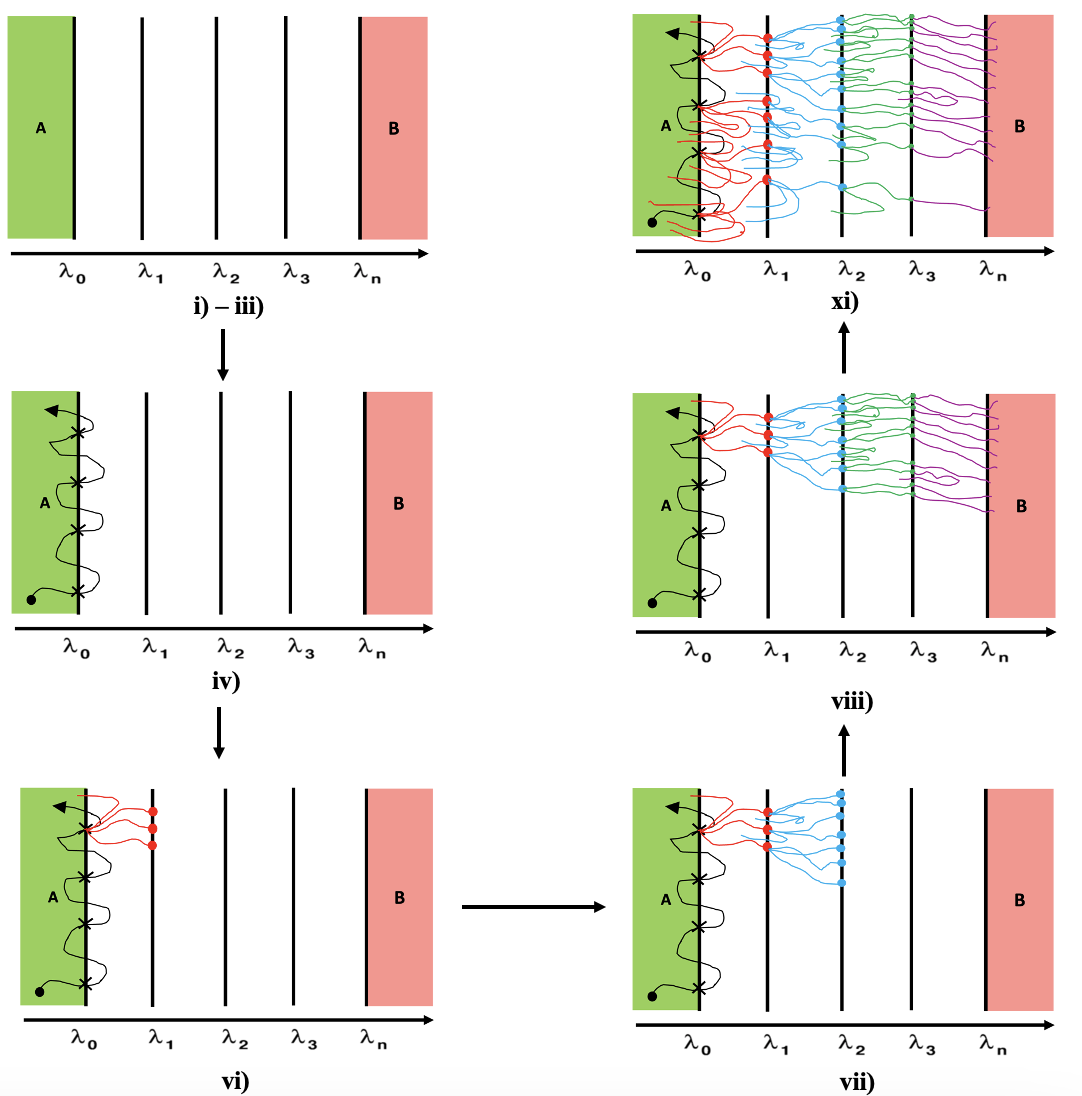}
    \caption{Schematic showing key steps for simulating abnormal trajectories using BG-FFS algorithm (refer to the points in Section \ref{sec_2.2})}
    \label{fig_2}
\end{figure}

\section{Materials and Methods}
\subsection{Overview of Key Steps}
\label{sec_2.1}

 Figure \ref{fig_1} provides an overview of the steps and methods utilized in this paper, with the steps described in subsequent sections.

\subsection{Step 1: Data Generation via Forward-flux Sampling}
\label{sec_2.2}

Path-sampling algorithms are Markov-Chain Monte-Carlo (MCMC)-based techniques, utilized routinely in molecular dynamics (MD) to analyze and simulate rare events from chemical reactants to products, including: crystal nucleation of hard spheres \cite{filion_crystal_2010} and sodium chloride \cite{jiang_forward_2018}, stochastic nonequilibrium systems \cite{allen_simulating_2006}, methane hydrate nucleation \cite{arjun_homogeneous_2023, bi_probing_2014}, and the like.  In previous research, as a first application of path-sampling algorithms for analyzing rare events for chemical process safety, \cite{moskowitz_understanding_2018} introduced transition-path sampling (TPS) [developed originally for MD by \cite{bolhuis_transition_2002, dellago_transition_1998, prigogine_transition_2002})], demonstrated on an exothermic CSTR and an air separation unit (ASU).  To overcome the computational limitations of TPS, \cite{sudarshan_understanding_2021} introduced forward-flux sampling (FFS) [developed originally for MD by \cite{allen_simulating_2006}]; i.e., from the same family of path-sampling algorithms as TPS, simulating rare un-postulated trajectories more-efficiently in a forward, piecewise manner, with the direct variant; i.e., DFFS introduced initially.  More recently, the branched-growth variant of FFS (i.e., BG-FFS) was utilized, generating trajectories more-suitable for committer analyses [conducted previously in MD by \cite{peters_obtaining_2006, borrero_reaction_2007}], resulting in improved, multivariate alarm systems for a P-only controlled exothermic CSTR \cite{sudarshan_multivariate_2023, sudarshan_alarm_2024} and a PID-controlled polystyrene CSTR \cite{sudarshan_path-sampling_2024}.  The steps involved in the BG-FFS algorithm, shown schematically in Figure \ref{fig_2}, include:

i) Define the initial desirable basin A and terminal undesirable basin B.\par
ii) Pick a suitable order parameter variable; i.e., $\lambda$; typically, this is a process variable that has a strong influence on the process dynamics; i.e., captures process deviations more-rapidly than other variables, and is not perturbed significantly using statistical noise; e.g., the reactor temperature.\par
iii) Based on the chosen $\lambda$, divide the space between the two basins into finite interfaces; i.e., $\lambda_{0},\lambda_{1},... \lambda_{n}$; where $\lambda_{0}$ and $\lambda_{n}$ represent the bounds for basins A and B.  Note that \textit{n} is the number of interfaces.\par
iv) Simulate a long initial trajectory that generates finite crossings across $\lambda_{0}$; if required, repeat this step for multiple trajectories to generate sufficient crossing points, with all process variables saved at every crossing point.\par
v) Compute the initial rate of transition across $\lambda_{0}$, $r_{0}$, as the total crossings divided by the total time spent in basin A by all the initial trajectories.\par
vi) Select a crossing point from among the saved crossings across $\lambda_{0}$ and simulate $m_{0}$ trajectories from that point, each of which continues until $\lambda_{1}$ is crossed.  Save the variables at all such crossing points.\par
vii) Simulate $m_{1}$ trajectories from every crossing point across $\lambda_{1}$ that generate crossing points across $\lambda_{2}$.  Save the variables at all such crossing points. \par
viii) Iterate step vii) for all subsequent interfaces till $\lambda_{n}$; i.e., simulate $m_{i}$ trajectories from all crossing points at $\lambda_{i}$ that continue until $\lambda_{i+1}$ is reached; save the variables at all such crossing points at $\lambda_{i+1}$; $\forall$ \textit{i} = 2, 3…\textit{n}-1.  \par
ix) Compute the overall transition probability of reaching basin B from basin A:
    \begin{equation}
    p_{\text{A} \rightarrow \text{B}} = \frac{N(\lambda_n \mid \lambda_0)}{\prod_{i=0}^{n-1} m_i}
    \label{eq_1}
    \end{equation}

    where $N(\lambda_n \mid \lambda_0)$ is the number of branches that reach basin B (i.e., from $\lambda_{n-1}$) and $\prod_{i=0}^{n-1} m_i$  are the total possible number of branches.\par
x) Compute the overall rate of transition, $r_{\text{A}\rightarrow\text{B}}$, as the product of $r_{0}$ and $p_{\text{A}\rightarrow\text{B}} (\lambda_{n} \mid \lambda_{0})$.\par
xi) Repeat steps iv) – x) for other crossing points at $\lambda_{0}$ and compute the average overall probability and rate of transition, i.e., $\overline{p}_{\text{A} \rightarrow \text{B}}$ and $\overline{r}_{\text{A} \rightarrow \text{B}}$. \par


\paragraph{}Note that every crossing point generated during the BG-FFS algorithm, with variables \textbf{x}, has an associated committer probability; i.e., $p_{\text{B}}(\textbf{x})$; i.e., the probability of a trajectory fired from that point reaches or “commits” to the terminal basin B.  The committer probabilities are computed recursively as \cite{borrero_reaction_2007}:

\begin{equation}
    p_j^i (\lambda_{i+1} \mid \lambda_i) = \frac{N_j^i}{m_i}
    \label{eq_2}
\end{equation}

\begin{equation}
\begin{split}
p_{\text{B}j}^i = p_j^i (\lambda_{i+1} \mid \lambda_i) \times \frac{\sum_{k=1}^{N_j^i} p_{\text{B}k}^{i+1}}{N_j^i} \\
= \frac{\sum_{k=1}^{N_j^i} p_{\text{B}k}^{i+1}}{m_i}, \quad i = n-1, n-2, \dots, 0
\end{split}
\label{eq_3}
\end{equation}

\paragraph{} where $p_{j}^i$ is the probability for a trajectory initiated from a point \textit{j} at $\lambda_{i}$ to reach the next interface; i.e., $\lambda_{i+1}$; $N_{j}^i$ is the number of successful trajectories reaching $\lambda_{i+1}$ from that point; $m_{i}$ is the total number of trajectories initiated from that point; and $p_{\text{B}j}$ is the committer probability for that point.  For example calculations, please refer to \cite{sudarshan_multivariate_2023, sudarshan_path-sampling_2024}.  Additionally, note that depending on the process parameter selected, initially as the response-action variable (i.e., a variable that is varied in real-time in response to alarms), the BG-FFS algorithm and $p_{\text{B}}$ calculations are repeated for multiple discrete values of  the response-action variable.  Hence, this makes the response-action variable \textit{discrete-valued}, whereas, the other variables saved during the BG-FFS algorithm and the estimated $p_{\text{B}}$ are \textit{continuous-valued}.

\subsection{Step 2: Data Preprocessing}
\label{sec_2.3}

As part of preprocessing, the $p_{\text{B}}$ – process variables data generated during the BG-FFS algorithm are filtered to remove outliers and structured in a clean, tabular format; i.e., with the process variables (e.g., temperature, concentration, and the like) as the input variables, and $p_{\text{B}}$ as the dependent variable.  Note that during BG-FFS, due to statistical noise-induced random perturbations, a wide distribution of $p_{\text{B}}$ is obtained for crossing points across each order parameter interface; i.e., $\lambda_i$.  Hence, to improve the predictions provided by the ML models, it is important to incorporate data filtering for these $p_{\text{B}}$.  Herein, simple filtering techniques are utilized to retain the $p_{\text{B}}$ centered around its mean; stated differently, only those data are retained that satisfy:

\begin{equation}
\overline{p}_{\text{B},i} - c_i \sigma_i \leq p_{\text{B},i} \leq \overline{p}_{\text{B},i} + c_i \sigma_i
\label{eq_4}
\end{equation}

\paragraph{} where $\overline{p}_{\text{B},i}$ and $\sigma_i$ are the mean and standard deviation of the $p_{\text{B}}$ for crossing points generated across $\lambda_i$; and $c_i$ is a filter factor, determined experimentally, such that neither too many nor too few data are filtered.  Hence, post preprocessing, the data are organized in a clean, tabular format, as shown in Table \ref{table_1}.  Note that additional preprocessing steps may be required depending on each ML algorithm.

\begin{table}[t]
\centering
\caption{Schematic for Tabular Data in our Analyses}  
\vspace{1em}
\renewcommand{\arraystretch}{1.5}  
\resizebox{\columnwidth}{!}{  
\begin{tabular}{c c c c c}  
\hline
\textbf{$\mathbf{\textit{p}_B}$} & \textbf{$\mathbf{\textit{X}_1}$} & \textbf{$\mathbf{\textit{X}_2}$} & \textbf{$\mathbf{\textit{X}_3}$} & \textbf{$\mathbf{\textit{X}_4}$} \\
\hline
$p_{\text{B},1}$ & $X_{1,1}$ & $X_{2,1}$ & $X_{3,1}$ & $X_{4,1}$ \\
$p_{\text{B},2}$ & $X_{1,2}$ & $X_{2,2}$ & $X_{3,2}$ & $X_{4,2}$ \\
$\vdots$  & $\vdots$  & $\vdots$  & $\vdots$  & $\vdots$  \\
$p_{\text{B},N_\text{samples}}$ & $X_{1,N_\text{samples}}$ & $X_{2,N_\text{samples}}$ & $X_{3,N_\text{samples}}$ & $X_{4,N_\text{samples}}$ \\
\hline
\end{tabular}
}  
\label{table_1}
\end{table}

\subsection{Step 3: Predictive Modeling via Machine Learning}
\label{sec_2.4}

Post data generation and preprocessing, using supervised machine learning, models are developed that predict $p_{\text{B}}$ for given process variables; i.e., a \textit{regression} problem is solved, given that $p_{\text{B}}$ is continuous-valued.  For each ML algorithm considered in this benchmark study, model development involves three steps:
\vspace{1em}

I) \textbf{Data Splitting}:  The preprocessed tabular data are divided into training and testing data, using randomized 70 $\%$-30$\%$ splits, as done routinely in practice \cite{bichri_investigating_2024, kahloot_algorithmic_2021, vrigazova_proportion_2021}.
\vspace{1em}

II) \textbf{Hyperparameter Optimization with Cross-Validation}:  Typically, ML models consist of two entities: hyperparameters to be optimized before training; and model training parameters learned during training.  The predictive performance of ML models is extremely sensitive to the choice of hyperparameters – hence, these need to be optimized carefully.  There are several open-source software packages available for hyperparameter optimization, including: Hyperopt \cite{bergstra_hyperopt_2015}, Optuna \cite{akiba_optuna_2019}, Ray tune \cite{liaw_tune_2018}, Optunity \cite{claesen_easy_2014}, and the like.  In this paper, the Optuna framework is chosen, utilizing a Bayesian optimization technique called a tree-structured parzen estimator; i.e.,  TPE \cite{bergstra_algorithms_2011, watanabe_tree-structured_2023}, to determine the optimum set of hyperparameters.  Additionally, in detailed benchmark studies comparing various optimization techniques and open-source frameworks, Optuna-TPE provided the most favorable performance and computation times \cite{motz_benchmarking_2022, shekhar_comparative_2022}.  Typically, the hyperparameter optimization process is carried out with $k$-folds cross validation:

\begin{figure*}[t]
    \centering
    \begin{subfigure}{0.5\textwidth}  
        \centering
        \includegraphics[width=\linewidth]{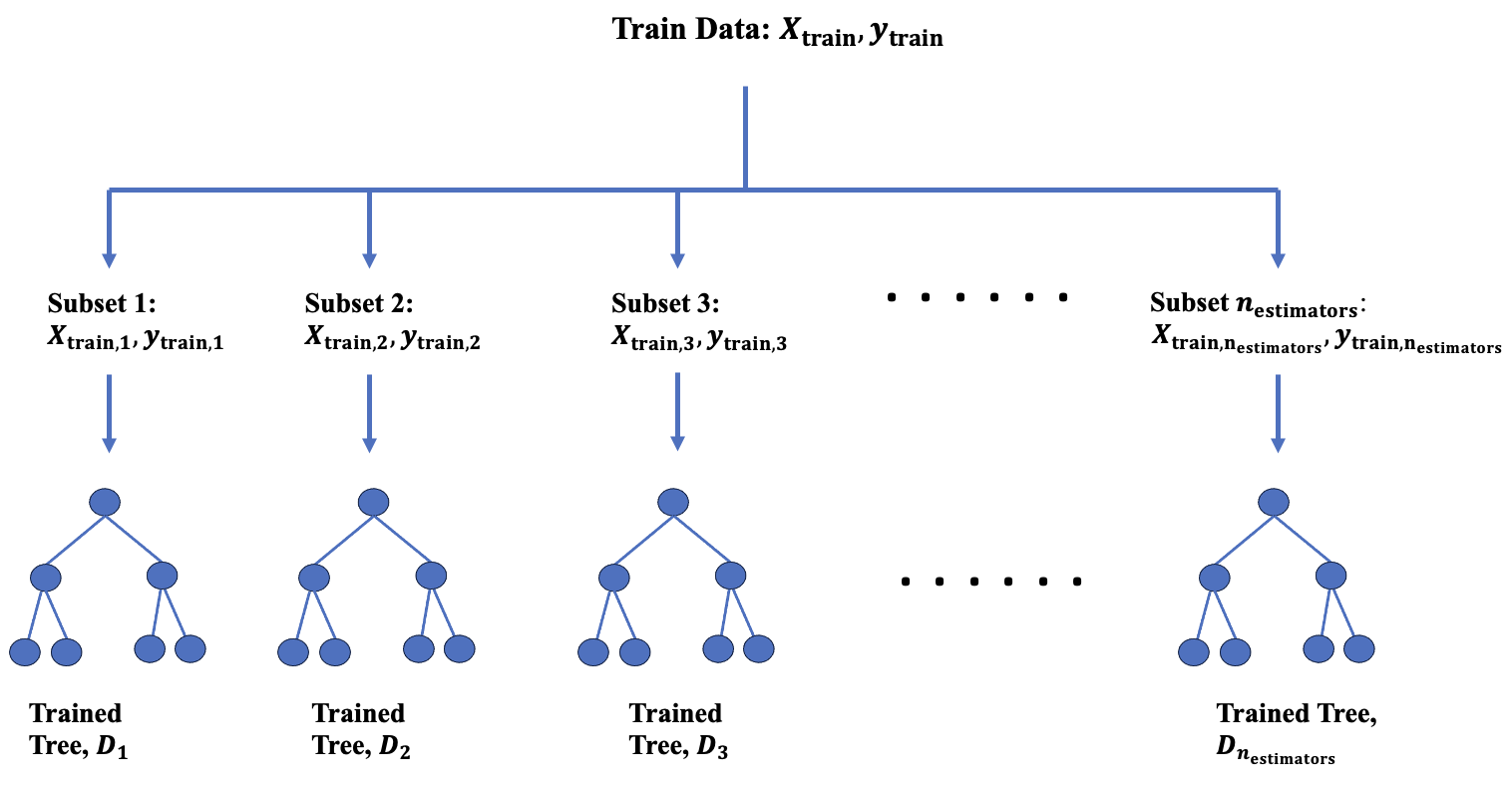}  
        \caption{$X_{\text{train}}$: Input variables for train data; $y_{\text{train}}$: Dependent variable for train data}
        \label{fig_3a}
    \end{subfigure}
    \hfill
    \begin{subfigure}{0.47\textwidth}  
        \centering
        \includegraphics[width=\linewidth]{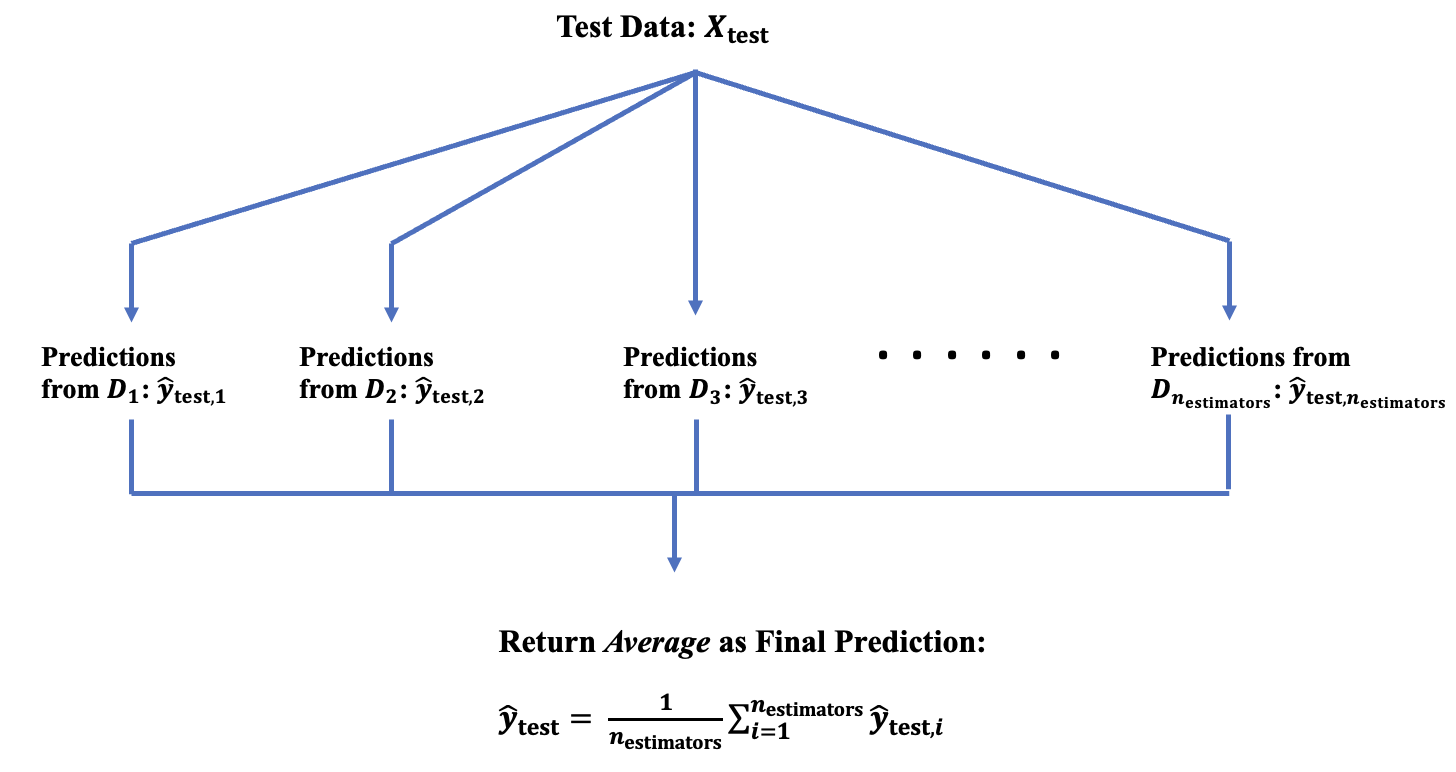}  
        \caption{$X_{\text{test}}$: Input variables for test data; $y_{\text{test,i}}$: Prediction for trained decision tree $i$, $D_i$, given $X_{\text{test}}$}
        \label{fig_3b}
    \end{subfigure}
    \caption{Schematic implementing key steps in RF for regression tasks, showing the: (a) Training phase; (b) Testing phase.}
    \label{fig_3}
\end{figure*}

a) Divide training data into $k$ sets (i.e., “folds”) randomly.  Herein, $k$ = 3.\par
b) Sample a combination of hyperparameters.\par
c) Set $i$ = 1.\par
d) Place set $i$ aside, and train the model using the remaining $k$ - 1 sets.  (When $k$ = 3, these are sets 2 and 3.)\par
e) Evaluate the performance of the trained model using set $i$ as the validation set and compute the validation score (e.g., \textit{RMSE} -- root-mean-squared-error). \par
f) When $i < k$, set $i$ = $i$ + 1.  Return to d).\par
g) When $i = k$, compute the average validation score.\par
h) Repeat steps b) – g).\par
i) Return the combination of hyperparameters that resulted in the maximum/minimum average evaluation score, depending on the chosen metric (e.g., return the combination that resulted in the minimum average \textit{RMSE}).\par
    
\vspace{1em}
III) \textbf{Model training with the Optimum Hyperparameters}: Post optimization, the ML model, with its optimum hyperparameters, is trained using the entire training data. 

\paragraph{} Note that eight ML algorithms of varying complexities are considered in this benchmark study.  These are described briefly: 

\vspace{1em}
1) \textbf{Linear SVR (Linear Support-Vector Regressor)}: An extension of the popular support-vector machines (SVM) algorithm developed originally for classification  problems [i.e., when the dependent variable is discrete-valued or categorical – e.g.,  “high”, “medium”, and “low”] \cite{cortes_support-vector_1995}, linear SVR is a parametric ML algorithm that involves training a linear model, referred to as \textit{hyperplane}, with a loss function that minimizes prediction error, also maintaining a tolerance margin; i.e., \textit{tube} \cite{drucker_support_1996}.  The parametric model for linear SVR is:

\begin{equation}
f(x) = w^T x + b
\label{eq_5}
\end{equation}

\paragraph{} where \textbf{x} is the vector of input features; \textbf{w} is the vector of coefficients (i.e., \textit{weights}), and \textit{b} is the \textit{bias} term, for each sample.  Next, the objective function to be minimized during training, and constraints are:

\begin{equation}
J(\mathbf{x}) = \frac{1}{2} \mathbf{w}^T \mathbf{w} + C \sum_{i=1}^{N_\text{samples}} (\xi_i + \xi_i^*)
\label{eq_6}
\end{equation}

\begin{equation}
\begin{aligned}
y_i - (\mathbf{w}^T \mathbf{x}_i + b) &\leq \epsilon + \xi_i \\
(\mathbf{w}^T \mathbf{x}_i + b) - y_i &\leq \epsilon + \xi_i^* \\
\xi_i, \xi_i^* &\geq 0
\end{aligned}
\label{eq_7}
\end{equation}

\paragraph{}where $C$ is the regularization term for mitigating overfitting; $N_{\text{samples}}$ is the number of samples; $\xi_i$ and $\xi_i^*$ are \textit{slack variables} (i.e., these define the penalty given to samples that violate the tolerance margin in the objective function); $\epsilon$ defines the tolerance margin.  Note that \textbf{w}, $b$, $\xi_i$, and $\xi_i^*$ are parameters learned during training, while $C$ and $\epsilon$ are hyperparameters.  For more details, please refer to \cite{drucker_support_1996}.  Additionally, note that Linear SVR requires additional preprocessing steps; i.e., categorical and discrete-valued input variables need to be transformed into integers.  Additionally, all input variables need to be scaled appropriately.
\vspace{1em}

2) \textbf{kNN (k-Nearest Neighbors)}:  kNN is a nonparametric supervised learning algorithm that estimates the relationship between the input features and output using the concept of \textit{nearest neighbors}.  Unlike most algorithms, kNN does not involve a training phase – for regression tasks, kNN estimates the output for an unknown sample (i.e., from the test data) by computing the average output of the $k$ known samples (i.e., from the training data) nearest to it.  kNN requires two key hyperparameters to be specified; i.e., the number of nearest-neighbors, $k$; and a distance metric for estimating the nearest samples – possible choices include: Euclidean, Manhattan, and Minkowski metrics \cite{danielsson_euclidean_1980,chen_comparative_2011,suwanda_analysis_2020}.  Similar to Linear SVR, kNN requires categorical and discrete-valued data to be transformed into integers, as well as appropriate scaling of input variables, given that kNN involves distance calculations that are sensitive to scaling.
\vspace{1em}

3) \textbf{RF (Random Forests)}: RF is a nonparametric learning algorithm \cite{breiman_random_2001} that involves training an ensemble of decision trees using \textit{bagging} (bootstrap aggregating); i.e., several decision trees (i.e., weak learners) are trained in parallel independently using different subsets of data that are sampled randomly (shown schematically in Figure \ref{fig_3a}).  Then, when predicting using test data, during regression tasks, the predicted output is computed as the average of the predictions provided by the trained trees (shown schematically in Figure \ref{fig_3b}).  Please refer to the Appendix, Section \ref{sec_A.4}, for a simple example of a decision tree.  Additionally, RF consists of several training parameters; e.g., optimal feature (i.e., input variable) for splitting, optimal split threshold for that feature, and the like.  Key hyperparameters include: \textit{n\_estimators}, (number of trees), \textit{max\_depth} (maximum depth for each tree), \textit{min\_samples\_split} (minimum number of samples required to split), and the like.  Please refer to \cite{breiman_random_2001} and \cite{scikit-learn_developers_randomforestregressor_2024} for more details regarding the training parameters, hyperparameters, and implementation of RF.  Additionally, note that RF requires categorical and discrete-valued input variables  to be transformed into integers, but does not require input features to be scaled.

\begin{figure}[t]
    \centering
    \begin{subfigure}{0.4\textwidth}  
        \centering
        \includegraphics[width=\linewidth]{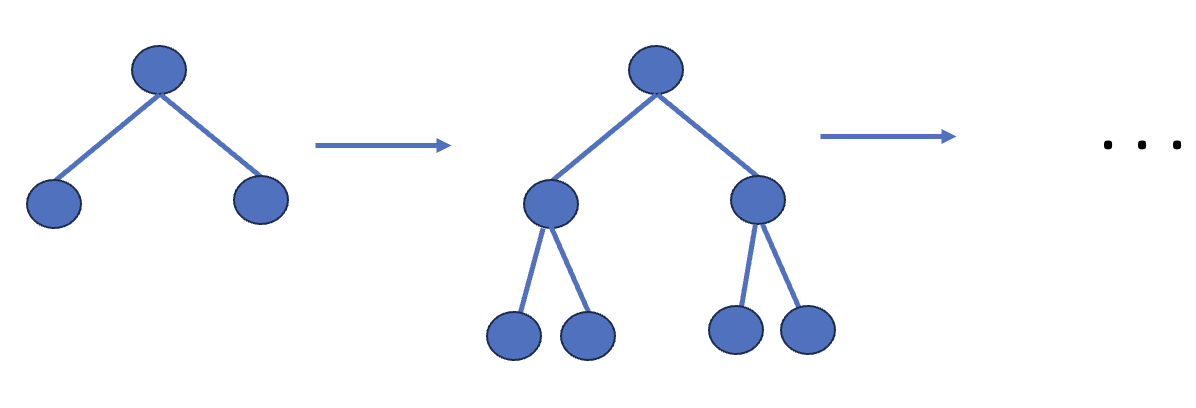}  
        \caption{}
        \label{fig_4a}
    \end{subfigure}
    \hfill
    \begin{subfigure}{0.45\textwidth}  
        \centering
        \includegraphics[width=\linewidth]{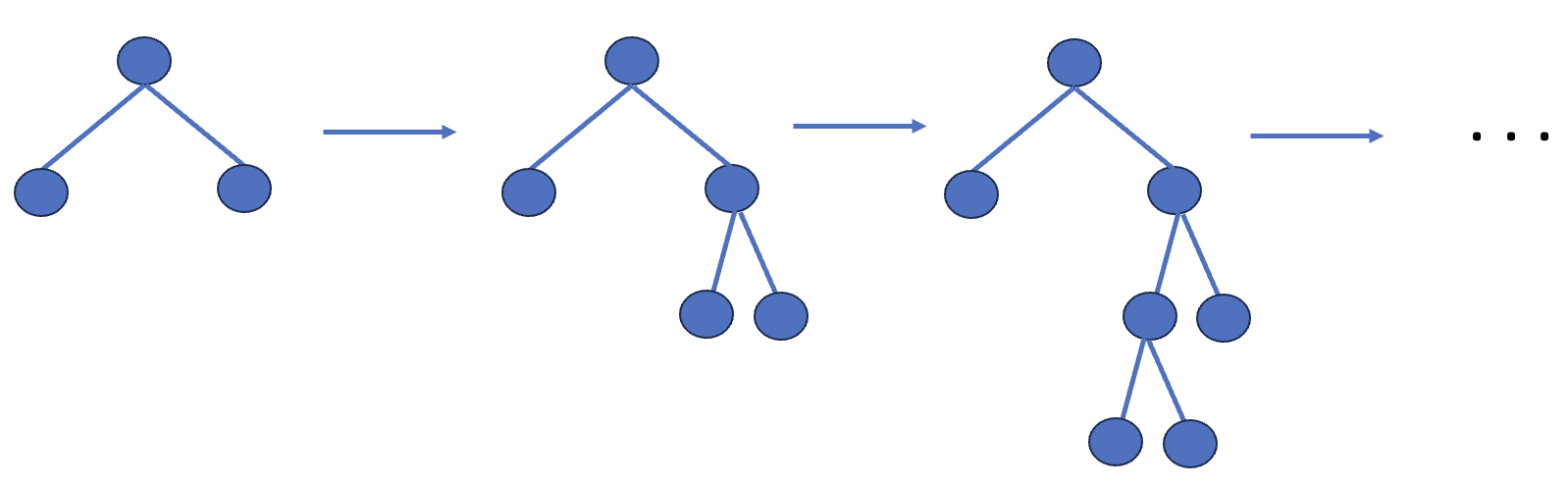}  
        \caption{}
        \label{fig_4b}
    \end{subfigure}
    \caption{(a) Level-wise growth strategy followed by XGBoost; (b) Leaf-wise growth strategy followed by LightGBM}
    \label{fig_4}
\end{figure}
\vspace{1em}
4) \textbf{XGBoost (eXtreme Gradient Boosting)}:  XGBoost is a nonparametric ensemble learning algorithm that belongs to the gradient-boosting family; i.e., several decision trees are trained sequentially, wherein, each newly-trained tree attempts to improve the predictions made by the previous trees (hence, the term “boosting”).  Additionally, the “eXtreme” component refers to additional regularization terms in the objective function to prevent overfitting; the “Gradient” term implies that new trees are trained using the gradients (i.e., first-order derivatives) and hessians (i.e., second-order derivatives) w.r.t the errors between the previous tree and the data.  Since its development by \cite{chen_xgboost_2016}, XGBoost has become a popular algorithm for tabular datasets, for regression and classification problems across several fields \cite{cerna_comparison_2020,li_gene_2019,ma_xgboost-based_2021,ogunleye_xgboost_2020}, including classification of rare events \cite{ashraf_identification_2023,wang_classification_2023}.\paragraph{} Note that in prior research, \cite{sudarshan_path-sampling_2024} developed improved dynamic bidirectional multivariate alarm systems for handling rare un-postulated abnormal events using XGBoost predictive models.  The training parameters for XGBoost are similar to those for RF and decision trees (e.g., optimal feature for splitting, optimal split threshold for the feature, and the like).  XGBoost also consists of several hyperparameters: \textit{n\_estimators}, \textit{max\_depth}, \textit{eta} (i.e., learning rate that scales the contribution of each tree), \textit{subsample} (i.e., specifies fraction of data used in training each tree – helps minimize overfitting by introducing randomness), \textit{reg\_alpha} (i.e., parameter for L1-norm regularization), \textit{reg\_lambda} (i.e., parameter for L2-norm regularization), and the like.  Note that some of these hyperparameters were optimized using Optuna appropriately by \cite{sudarshan_path-sampling_2024} while developing XGBoost models.  For more details on the algorithm and hyperparameters, please refer to \cite{chen_xgboost_2016} and \cite{xgboost_developers_xgboost_2023}.

\vspace{1em}
5) \textbf{LightGBM (Light Gradient-Boosting Machines)}:  Like XGBoost, LightGBM is a ML algorithm by \cite{ke_lightgbm_2017} that belongs to the gradient-boosting family.  The key difference between XGBoost and LightGBM is: trees in XGBoost follow a \textit{level}-wise growth strategy (i.e., two resulting nodes at each level are split simultaneously), whereas trees in LightGBM follow a \textit{leaf}-wise growth strategy (i.e., only one of the nodes, chosen optimally, is split further at each level \cite{liang_predicting_2020}), potentially reducing model development times – these are shown schematically in Figure \ref{fig_4}.  Note that the training parameters and hyperparameters for LightGBM are similar to those for XGBoost.  For more details, please refer to \cite{ke_lightgbm_2017}.

\vspace{1em}
6) \textbf{CatBoost (Categorical Boosting)}:  CatBoost is another ML algorithm belonging to the gradient-boosting family, developed by \cite{prokhorenkova_catboost_2018}.  Compared to XGBoost and LightGBM, CatBoost implements two improvements to mitigate overfitting: i) Ability to handle data with categorical input features more-efficiently by calculating \textit{ordered target statistics}; ii) Implementing \textit{ordered boosting} – as per \cite{prokhorenkova_catboost_2018}, gradient-boosting frameworks developed previously suffer from prediction shift; i.e., typically, each tree in the ensemble is trained using the entire training data, leading potentially to overfitting.  In ordered boosting, each tree is trained using \textit{random permutation sets} of the training data, using only the data before each example in the permutation set – ensuring improved robustness of models in the face of unseen data.  Several articles have compared the performance of XGBoost, LightGBM, and CatBoost across different applications for: home-credit dataset \cite{daoud_comparison_2019}, insurance claims \cite{so_enhanced_2024}, medicare fraud detection \cite{hancock_performance_2020}, to name a few.  Note that the training parameters and hyperparameters for CatBoost are similar to XGBoost and LightGBM.  For more details, please refer to \cite{prokhorenkova_catboost_2018}.
\vspace{1em}

7) \textbf{DNN (Dense Neural Network)}:  Deep learning models based on Artificial Neural Networks (ANNs) have become popular across several fields, most notably for computer-vision applications such as image recognition \cite{traore_deep_2018} and video encoding \cite{ma_image_2020}; natural-language processing (NLP), such as dialogue summarization \cite{chen_dialogsum_2021}; machine translation \cite{singh_machine_2017}; sentiment analysis \cite{dos_santos_deep_2014}, and the like.  The basic building blocks of ANNs are referred to as \textit{perceptrons}, developed by \cite{rosenblatt_perceptron_1958}.  A single perceptron model consists of:  i) A Linear model, $f(\textbf{x})$, comprising of weights \textbf{w}, and bias $b$ (shown previously in Eq. (\ref{eq_5}), where \textbf{x} is the vector of input features, and; ii) An activation function, $g(f(\textbf{x}))$, applied to the output of the linear model.  Choices for activation function include: rectified linear unit (i.e., \textit{ReLU}) \cite{nair_rectified_2010}, hyperbolic tangent (i.e., \textit{tanh}), sigmoid (suitable for binary classification tasks), and the like.  DNNs consist of multiple layers of several perceptrons, with the DNN referred to as a deep DNN when there are two or more hidden layers (i.e., the layers between the input and the output).  DNN also require a loss function to be specified (for regression tasks, this is typically the mean-squared-error; i.e., \textit{MSE}), with the weights and biases optimized during training through \textit{back-propagation} over several training \textit{epochs}.  Additionally, several optimization routines exist for DNN training: gradient-descent, stochastic gradient-descent \cite{ruder_overview_2017}, Adam (Adaptive Moment Estimation) \cite{kingma_adam_2017}, and the like.  For detailed explanations, please refer to \cite{zou_overview_2009} and \cite{hassoun_fundamentals_1995}.  Additionally, similar to Linear SVR and kNN, DNNs require categorical and discrete-valued input variables to be transformed into integers, as well as appropriate scaling of input variables.
\vspace{1em}

8) \textbf{TabNet (Tabular Networks)}:  Given that gradient-boosting frameworks regularly outperform DNNs on tabular datasets, as observed over several benchmark studies \cite{borisov_deep_2022,grinsztajn_why_2022,shwartz-ziv_tabular_2022,uddin_confirming_2024}, efforts have been made to develop novel neural network-based architectures, specifically for tabular data.  One, developed by \cite{arik_tabnet_2020}, is TabNet.  Compared to DNNs, TabNet utilizes an \textit{attention mechanism} (i.e., an attentive transformer layer that assigns weights to different input features, with important features weighted more heavily) to select the input features most influential for predictions.  Note that this feature-selection capability is inspired from tree-based, gradient-boosting models, wherein, each tree splits data related to the most important features.  Additionally, unlike DNNs, TabNet relies less on data preprocessing, with categorical input features processed internally using \textit{embedding} layers.  More recently, several research articles have considered TabNet for applications such as: rainfall forecasting, \cite{yan_rainfall_2021}, electric load forecasting \cite{borghini_short_2021}, diabetes classification \cite{joseph_explainable_2022}, insurance claims prediction \cite{mcdonnell_deep_2023}, to name a few.  For more details regarding the architecture, parameters, and hyperparameters, please refer to \cite{arik_tabnet_2020}.

\subsection{Step 4: Reporting Key Evaluation Metrics}
\label{sec_2.5}
Post development, for each dataset, all ML models are evaluated comprehensively and holistically across three key domains:

\vspace{1em}
A) \textbf{Model Accuracy on Test Data}: The accuracy of the trained ML model on the test data is evaluated using:

\begin{equation}
\begin{split}
\textit{RMSE}\, (\text{metric}_1) &= \textit{Root Mean Squared Error} \\
&= \sqrt{\frac{\sum_{i=1}^{N_{\text{samples}}} \left(p_{\text{B,test}}(i) - \hat{p}_{\text{B,test}}(i)\right)^2}{N_{\text{samples}}}}
\end{split}
\label{eq_8}
\end{equation}

\vspace{1em}
B) \textbf{Computational Efficiency}: Four clock-times are recorded to evaluate computational costs: 

i) $t_\text{hyper} \, (\text{metric}_2)$:  Time recorded for hyperparameter optimization (i.e., Step II, Section \ref{sec_2.4}).\par
ii) $t_\text{train} \, (\text{metric}_3)$ : Time recorded for model training (i.e., Step III, Section \ref{sec_2.4}).\par
iii) $t_\text{test} \, (\text{metric}_4)$: Time recorded for model testing; i.e., time taken to generate committer probability predictions for the test data.\par
iv) $t_\text{deploy} \, (\text{metric}_5)$: Time recorded for model deployment; i.e., time taken by each ML model to generate new committer probability predictions on-line for a new dynamic simulation (i.e., the number of dynamic simulations, $N_\text{sim}$ = 1), $t_\text{sim}$, time for a simulation, and model call frequency, $call_\text{freq}$ (i.e., frequency an ML model is called to generate new predictions on-line); e.g., $call_\text{freq}$ = 30 indicates the ML model is called once every 30 time-steps on-line to generate fresh predictions.  Note that small $call_\text{freq}$ leads to more-frequent on-line predictions of $p_{\text{B}}$ at excessive computational costs.\par
\vspace{1em}
C) \textbf{Alarm-system Efficiency}: To evaluate the impact of each ML model on efficiency of alarm systems, first, we need to define a specific alarm system.  For all datasets utilized in this benchmark study, a 2-level alarm system is assumed (i.e., number of alarm levels, $n_{\text{levels}}$ = 2) based on $p_{\text{B}}$ limits (i.e., an alarm at level $k$ is activated when the real-time $p_{\text{B}}$ predicted by a ML model crosses the $p_{\text{B}}$ limit defined at that level): for this benchmark study, these limits were set at $p_{\text{B},1}$ = 0.2, and, $p_{\text{B},2}$ = 0.5; where $p_{\text{B},1}$ and $p_{\text{B},2}$  are the $p_{\text{B}}$ limits defined for levels 1 and 2.  Then, the theoretical performance of each alarm level is computed using $p_{k,\text{theoretical}}$  – defined as the theoretical probability with which, alarms, active at the current level $k$, reach the next level.  Table \ref{table_2} shows the 2-level alarms specified and the associated $p_{k,\text{theoretical}}$ values.  For instance,  $p_{1,\text{theoretical}}$ = $p_{\text{B},1}⁄p_{\text{B},2}$ = $0.2/0.5$ = 0.4.  And, $p_{2,\text{theoretical}}$  = $p_{\text{B},2}/1.0$ = 0.5, given that level 2 is the last alarm-level before the undesirable region is reached (where $p_{\text{B}}$ = 1.0).  

\begin{table}
\centering
\caption{2-level Alarms Specified and Associated $p_{k,\text{theoretical}}$}
\vspace{1em}
\renewcommand{\arraystretch}{1.5}
\small  
\begin{tabular}{c c c}  
\hline
\textbf{Level No.} & \boldmath$p_{\text{B},k}$ & \boldmath$p_{k,\text{theoretical}}$ \\
\textbf{(i.e., k)} &  &  \\
\hline
1 & 0.2 & 0.4 \\
2 & 0.5 & 0.5 \\
\hline
\end{tabular}
\label{table_2}
\end{table}

\paragraph{} Next, for each ML algorithm, the performance of the alarm system is measured using $p_{k,\text{measured}}$ – defined as the probability for alarms active at the current level $k$ of reaching the next level (i.e., $k+1$) measured over several dynamic simulations, using real-time ML model predictions on-line, computed as:

\begin{equation}
p_{k,\text{measured}} = \frac{n_{\text{alarms},k \rightarrow k+1}}{n_{\text{alarms},k}}
\label{eq_9}
\end{equation}

where $n_{\text{alarms},k}$ is the number of alarms active at level $k$; $n_{\text{alarms},k \rightarrow k+1}$ is the number of alarms at level $k$ that are active when the process reaches level $k+1$.  Given $p_{k,\text{theoretical}}$ and $p_{k,\text{measured}}$, two alarm metrics are proposed to evaluate the alarm-system efficiency; i.e., absolute probability difference ($\Delta p$), and \textit{Total Alarms}:

\begin{equation}
\Delta p \, (\text{metric}_6) = \sum_{k=1}^{n_{\text{levels}}} k \left| p_{k,\text{theoretical}} - p_{k,\text{measured}} \right|
\label{eq_10}
\end{equation}

\begin{equation}
\textit{Total Alarms} \, (\text{metric}_7) = n_{\text{alarms}} = \sum_{k=1}^{n_{\text{levels}}} n_{\text{alarms},k}
\label{eq_11}
\end{equation}

\paragraph{} Eq. (\ref{eq_10}) accounts for both \textit{false positive rates} (i.e., fewer than expected alarms at the current level remain active when the process reaches the next level) and \textit{false negative rates} (i.e., more than expected alarms at the current level remain active when the process reaches the next level).  Additionally, as per Eq. (\ref{eq_10}), differences in higher-level alarms are penalized more heavily.  $\Delta p$ and \textit{Total Alarms} are recorded over several dynamic simulations (e.g., $N_\text{sim}$ ~ 50), with simulation time, $t_\text{sim}$, and model call-frequency, $call_\text{freq}$.  While recording these metrics, response actions are not included, given that $p_{k,\text{theoretical}}$ estimates do not account for changes in process dynamics when response actions are activated.  Additionally, to ensure consistency, while recording $t_{\text{deploy}}$, $\Delta p$, and \textit{Total Alarms}, for each dynamic simulation, a random seed number is utilized across all models (e.g., 50 random seed numbers are utilized for $N_\text{sim}$ = 50).  Note that using a random seed number ensures that the same sequence of statistical noise samples is generated for a dynamic simulation – enabling consistent comparison among ML algorithms.

\subsection{Step 5: Determine Optimal ML Model}
\label{sec_2.6}
Given the evaluation metrics defined in Section \ref{sec_2.5}, for our benchmark analyses, a weighted cost function to be minimized is proposed:

\begin{equation}
\textit{Cost} = \sum_{i=1}^{n_{\text{metrics}}} \left( a_i \right) \left(\text{metric}_{i,\text{scaled}}\right)
\label{eq_12}
\end{equation}

\begin{equation}
\begin{split}
[a_1, a_2, a_3, a_4, a_5, a_6, a_7] &= \\
[0.125, 0.05, 0.05, 0.05, 0.125, 0.3, 0.3]
\end{split}
\label{eq_13}
\end{equation}

\begin{table*}[h]
\centering
\caption{Response-action Variables and Terminal Regions for Various Datasets}  
\renewcommand{\arraystretch}{1.5}  
\resizebox{\textwidth}{!}{  
\begin{tabular}{c c c c}  
\hline
\textbf{Dataset} & \textbf{Response-action Variable} & \textbf{Terminal Region} & \textbf{Discrete Values Considered for Response-action Variable} \\
\hline
II & $q_{\text{i}}$ & Unsafe & [0.0875, 0.09, 0.095, 0.0975, 0.1, 0.1025, 0.105, 0.1075] \\
III & $q_{\text{i}}$ & Unreliable & [0.08, 0.085, 0.09, 0.095, 0.1, 0.105, 0.11] \\
IV & $q_{\text{m}}$ & Unsafe & [0.37, 0.375, 0.3775, 0.38, 0.385, 0.39, 0.4, 0.405] \\
V & $q_{\text{m}}$ & Unreliable & [0.375, 0.3775, 0.38, 0.385, 0.39, 0.4, 0.405] \\
\hline
\end{tabular}
}  
\label{table_3}
\end{table*}

where $n_\text{metrics}$ is the total number of evaluation metrics; i.e., $n_\text{metrics}$ = 7; $a_i$ is the weighting coefficient for $\text{metric}_i$.  To ensure consistency in scaling, each evaluation metric in Eq. (\ref{eq_12}) is scaled by its maximum value obtained across all ML models.  As per Eq.(\ref{eq_13}), the coefficients are weighed such that primary importance is allocated to alarm-system efficiency (i.e., $\Delta p$, and \textit{Total Alarms}), followed by model accuracy and model deployment, with clock-times concerning hyperparameter optimization, model training, and model testing weighed least.  \textbf{Hence, for each dataset, the optimal ML model has the lowest \textit{Cost}}.

\section{Results and Discussions}

This section provides results for the benchmark analysis obtained for each of the five datasets.  Note that for all evaluation metrics and $Cost$ estimates, lower values indicate better performance.

\begin{figure}[h]
    \centering
    \begin{subfigure}{0.45\textwidth}  
        \centering
        \includegraphics[width=\linewidth]{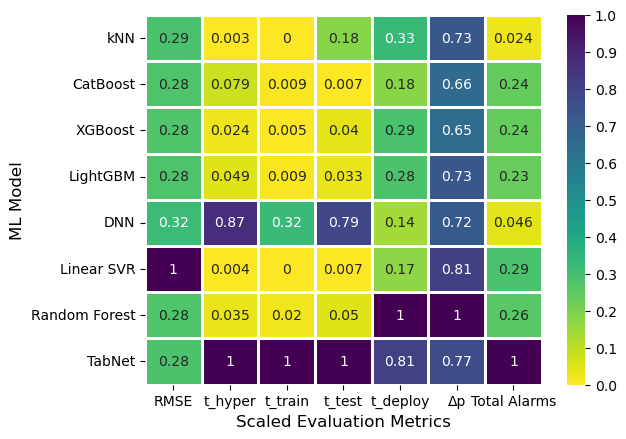}  
        \caption{}
        \label{fig_5a}
    \end{subfigure}
    \hfill
    \begin{subfigure}{0.4\textwidth}  
        \centering
        \includegraphics[width=\linewidth]{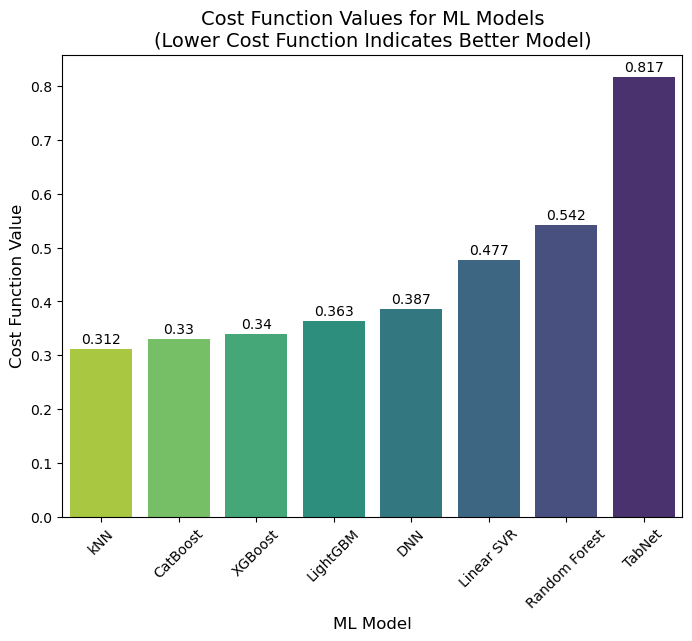}  
        \caption{}
        \label{fig_5b}
    \end{subfigure}
    \caption{For Dataset 1, (a) Heatmap showing the scaled evaluation metrics; (b) Cost computed for all ML models }
    \label{fig_5}
\end{figure}

\subsection{Dataset I (PI-Controlled Exothermic CSTR)}
\label{sec_3.1}
For the exothermic CSTR with abnormal transitions towards the unreliable region (see the process model summarized in Appendix \ref{sec_A.1}), the discrete values considered for the response-action variable, residence time, $\tau$ \cite{sudarshan_multivariate_2023}, are:  

\begin{equation}
\tau \in \{0.53, 0.54, 0.55, 0.56, 0.57, 0.58, 0.59\}
\label{eq_14}
\end{equation}

Note that all other process parameters remain constant (see Table \ref{table_A1} in the Appendix).  Additionally, for recording metrics concerning alarm-system efficiency, $N_\text{sim}$ = 50; $t_\text{sim}$ = 30,000 mins; $call_\text{freq}$ = 200; and $\tau$ = 0.53 min.

\begin{figure*}[h]
    \centering
    \begin{subfigure}{0.42\textwidth}  
        \centering
        \includegraphics[width=\linewidth]{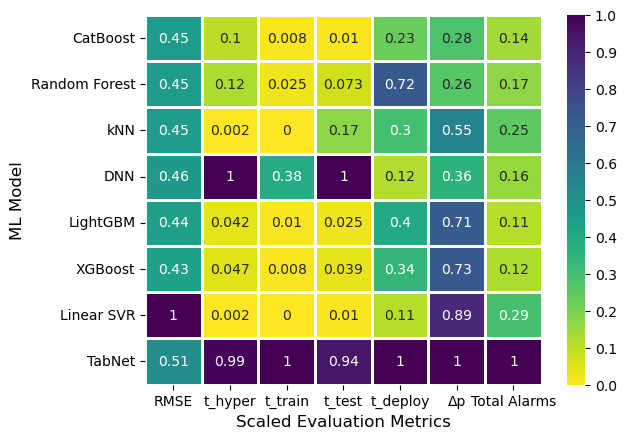}  
        \caption{}
        \label{fig_6a}
    \end{subfigure}
    \hfill
    \begin{subfigure}{0.42\textwidth}  
        \centering
        \includegraphics[width=\linewidth]{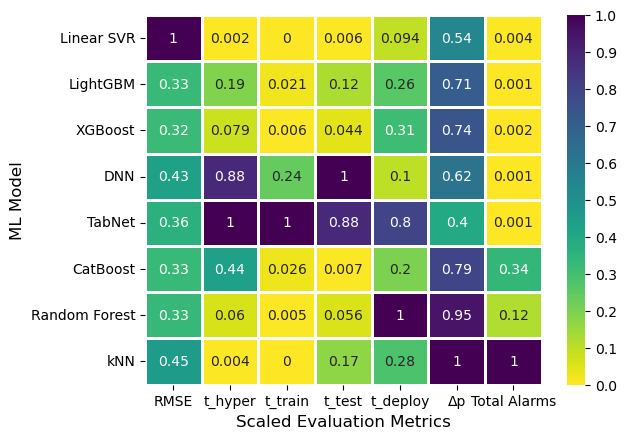}  
        \caption{}
        \label{fig_6b}
    \end{subfigure}
    \vspace{1em}  
    \begin{subfigure}{0.42\textwidth}  
        \centering
        \includegraphics[width=\linewidth]{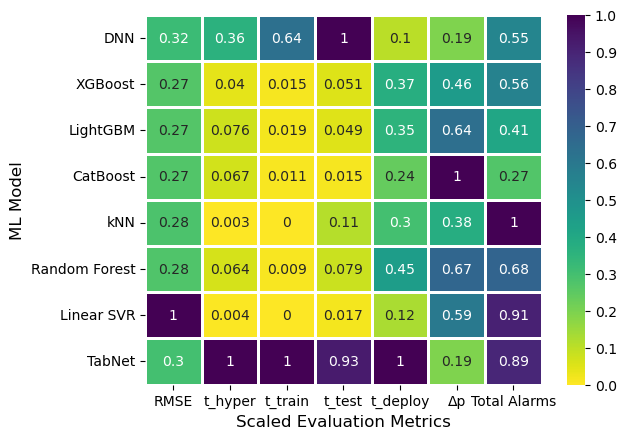}  
        \caption{}
        \label{fig_6c}
    \end{subfigure}
    \hfill
    \begin{subfigure}{0.42\textwidth}  
        \centering
        \includegraphics[width=\linewidth]{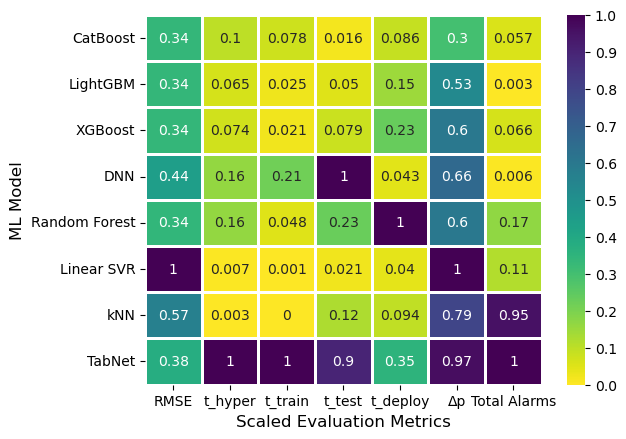}  
        \caption{}
        \label{fig_6d}
    \end{subfigure}
    \caption{Heatmaps showing scaled evaluation metrics for datasets (a) II; (b) III; (c) IV; (d) V.}
    \label{fig_6}
\end{figure*}

\begin{figure*}[h]
    \centering
    \begin{subfigure}{0.45\textwidth}  
        \centering
        \includegraphics[width=\linewidth]{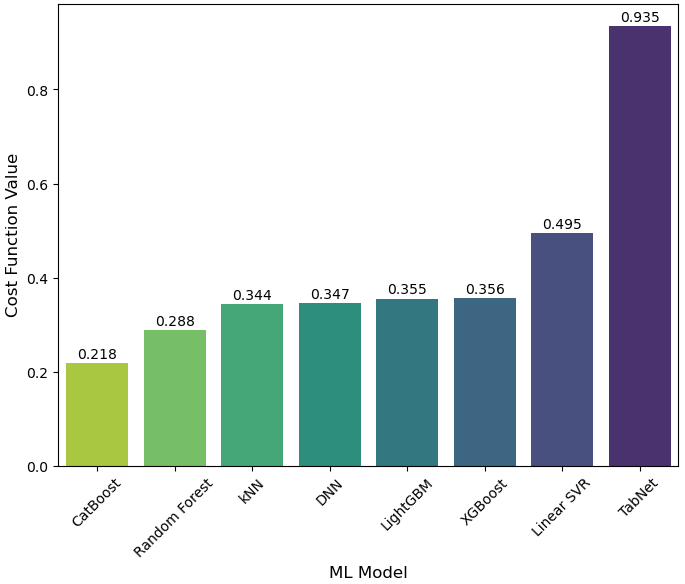}  
        \caption{}
        \label{fig_7a}
    \end{subfigure}
    \hfill
    \begin{subfigure}{0.45\textwidth}  
        \centering
        \includegraphics[width=\linewidth]{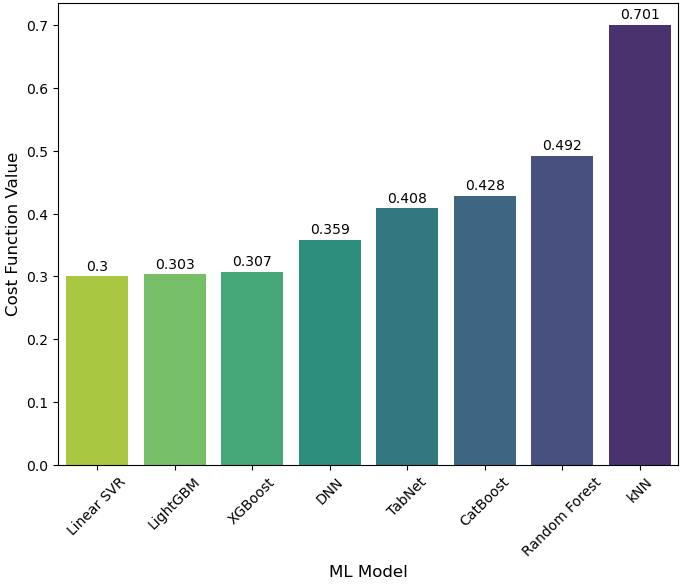}  
        \caption{}
        \label{fig_7b}
    \end{subfigure}
    \vspace{1em}  
    \begin{subfigure}{0.45\textwidth}  
        \centering
        \includegraphics[width=\linewidth]{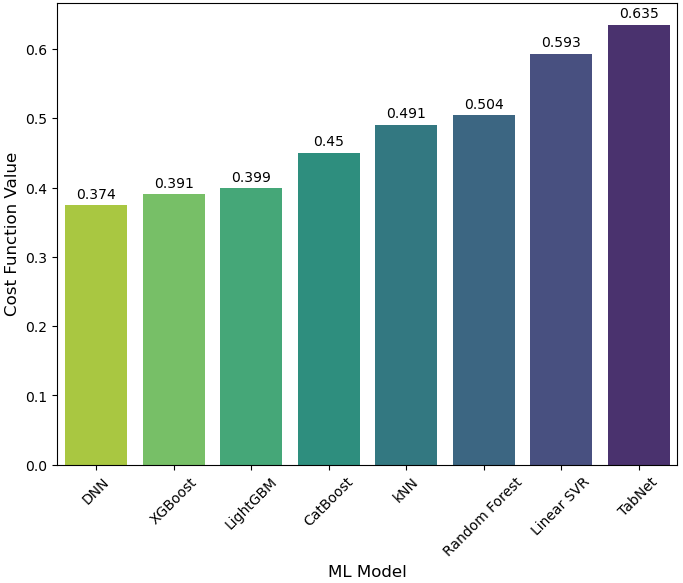}  
        \caption{}
        \label{fig_7c}
    \end{subfigure}
    \hfill
    \begin{subfigure}{0.45\textwidth}  
        \centering
        \includegraphics[width=\linewidth]{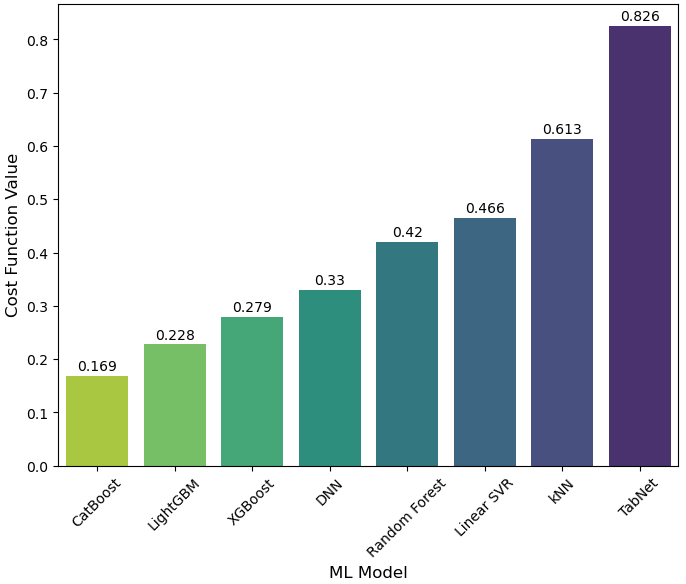}  
        \caption{}
        \label{fig_7d}
    \end{subfigure}
    \caption{Cost computed for all ML models for datasets (a) II; (b) III; (c) IV; (d) V.}
    \label{fig_7}
\end{figure*}

\paragraph{}Figure \ref{fig_5a} shows heatmaps for scaled evaluation metrics recorded for dataset I for the eight ML algorithms.  For each metric, lighter colors indicate better performance.  Most algorithms show comparable performance for model accuracy (i.e., $\textit{RMSE}_{\text{scaled}} \sim 0.28-0.29$), with Linear SVR having the worst model accuracy, and DNN having slightly-higher \textit{RMSE}.  Note that high \textit{RMSE} indicates less-accurate on-line $p_{\text{B}}$ predictions, potentially contributing to increased false and missed alarms.  With a relatively-simple development process, Linear SVR compensates slightly with higher computational efficiency, followed by kNN and gradient-boosting frameworks, with DNN having low costs for deployment despite higher costs for training, hyperparameter optimization, and testing.  For alarm-system efficiency, kNN and DNN perform significantly better – note that higher $\Delta p$ and \textit{Total Alarms} indicate increased false alarms and missed alarms, potentially resulting in alarm flooding (i.e., with the number of alarms significantly greater than operators can handle, leading to operator distraction and missed alarms).  Despite much promise for tabular datasets, TabNet underperforms significantly in most evaluation metrics.  Additionally, despite lower model development costs (i.e., $t_{\text{hyper}}$ and $t_{\text{train}}$), RF records high costs for model deployment (i.e., $t_{\text{deploy}}$) – potentially resulting in a lag between on-line process variable measurements and $p_\text{B}$ predictions. Figure \ref{fig_5b} shows the Cost computed for each ML algorithm.  Given the weights defined in Eq. (\ref{eq_13}), for Dataset I, kNN is observed to have the lowest cost, and is ranked as the most-optimal ML algorithm, followed closely by CatBoost and XGBoost, with TabNet ranking last.

\begin{figure}[h]
    \centering
    \includegraphics[width=0.9\linewidth]{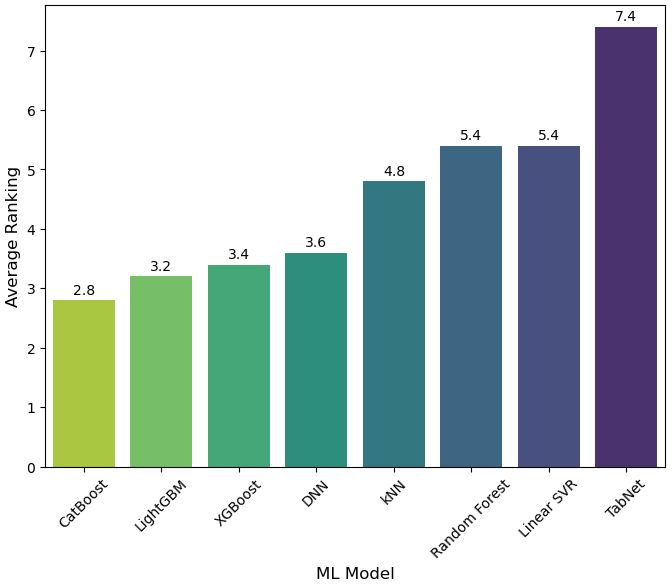}
    \caption{Average local rankings computed for models across all datasets for weights in Eq. (\ref{eq_13})}
    \label{fig_8}
\end{figure}

\subsection{Datasets II – V (PID-Controlled Polystyrene CSTR)}
\label{sec_3.2}
For the PID-controlled polystyrene CSTR (for the process model summarized in Appendix \ref{sec_A.2}), Table \ref{table_3} shows the specifications; i.e., the response-action variable, terminal region, and discrete values considered for the response-action variables.  For each dataset, all other process parameters remain constant (see Table \ref{table_A2}).  Note that for datasets II and III, $q_{\text{m}}$ = 0.4; and, for datasets IV and V, $q_{\text{i}}$ = 0.1.  Additionally, for metrics concerning alarm efficiency, $N_{\text{sim}}$ = 50; $t_{\text{sim}}$ = 150; $call_\text{freq}$ = 30; $q_{\text{i}}$ = 0.1; and $q_{\text{m}}$ = 0.4. 

Figure \ref{fig_6} a-d show heatmaps for scaled evaluation metrics for datasets II – V.  For model accuracy, gradient-boosting algorithms achieve stronger predictive performance, with Linear SVR scoring the lowest \textit{RMSE}, given its low-complexity.  More-complex DNN and TabNet algoritms do not justify their \textit{RMSE} scores.  For computational efficiency, the less-complex algorithms, Linear SVR and kNN offer fast computational times, followed by the gradient boosting algorithms, with DNN and TabNet having lowest computational efficiency (except fast model deployment for DNN, consistent with that observed for dataset I).  Additionally, RF records high costs for deployment, despite relatively-lower costs for model development, as observed for dataset I.  For alarm efficiency, performance varies across the datasets, with best being CatBoost and RF for dataset II; Linear SVR and LightGBM for dataset III; DNN, XGBoost, and LightGBM for dataset IV; CatBoost and LightGBM for dataset V.  Note – TabNet performs poorly uniformly.

\paragraph{} For datasets II-V, Figure \ref{fig_7} a-d show the $Cost$ for all ML algorithms, given the weights defined in Eq.(\ref{eq_13}).  CatBoost ranks as the most-optimal ML model for datasets II and V, with DNN the most-optimal for dataset IV.  Note that despite poor model accuracy, Linear SVR compensates by having improved computational and alarm efficiencies, thereby, unexpectedly ranking as the most-optimal ML algorithm for dataset III.  With the exception of dataset III, TabNet is the least-optimal model.

\subsection{Average Rankings Across All Datasets for Weights in Eq. (\ref{eq_13})}
\label{sec_3.3}
For all ML models, Figure \ref{fig_8} shows the average (mean) rankings across all five datasets, using the weighting coefficients in Eq. (\ref{eq_13}).  Clearly, the gradient-boosting frameworks achieve favorable rankings, with CatBoost achieving the highest ranking, followed by LightGBM and XGBoost, with RF, Linear SVR and TabNet recording the lowest rankings.
\begin{figure}[h]
    \centering
    \includegraphics[width=0.95\linewidth]{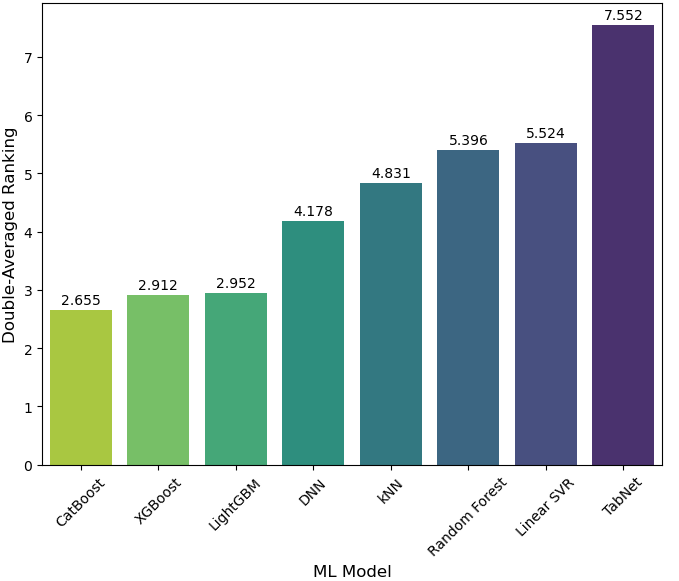}
    \caption{Double-averaged, global ranking for 500 combinations of weighting coefficients across all datasets}
    \label{fig_9}
\end{figure} 

\subsection{Double-Averaged Global Ranking Across All Datasets and 500 Weight Combinations}
\label{sec_3.4}
Note that the $Cost$ results in Sections \ref{sec_3.1}-\ref{sec_3.3} (i.e., Figures \ref{fig_5b}, \ref{fig_7}, and \ref{fig_8}) are for the single combination of weighting coefficients in Eq. (\ref{eq_13}).  For more-comprehensive analyses, several combinations of weighting coefficients are justified.  Herein, 500 different combinations are considered, with each combination sampled randomly from uniform distributions, given upper and lower bounds:

\begin{equation}
\begin{aligned}
0.1 &\leq a_1 \leq 0.2 \\
0.05 &\leq a_2 \leq 0.1 \\
0.05 &\leq a_3 \leq 0.1 \\
0.05 &\leq a_4 \leq 0.1 \\
0.1 &\leq a_5 \leq 0.2 \\
0.3 &\leq a_6 \leq 4 \\
0.3 &\leq a_7 \leq 4
\end{aligned}
\label{eq_15}
\end{equation}

Note that while the choice of distribution is arbitrary, with the normal distribution and others possible, these weighting coefficient bounds emphasize alarm-system efficiency over other metrics.
\paragraph{} As shown in Figure \ref{fig_9}, the use of these randomly-sampled weighting coefficients provides a similar ranking.  CatBoost achieves the highest global ranking, followed by XGBoost, LightGBM, and DNN, with  TabNet achieving the lowest ranking.  The increased ranking of XGBoost may warrant further consideration.

\section{Conclusions}

In previous research, using path-sampling (i.e., BG-FFS) and non-parametric machine learning, dynamic, bidirectional multivariate alarm systems were developed for rare un-postulated abnormal movements away from normal operating regions, demonstrated successfully on a PID-controlled polymerization CSTR.  However, in predictive modeling, only one ML algorithm was explored; i.e., XGBoost.
\paragraph{} Herein, a comprehensive framework is developed for benchmark analyses to explore optimal ML algorithms of varying complexities, for enhancing predictions of rare abnormal events using chemical process models.  For evaluation, several metrics are considered, permitting balances between model accuracy, and computational and alarm-system efficiencies, with more preference given to alarm-system efficiencies.  We conclude that the evaluation presented, placing emphasis on the number and efficiency of alarms activated, significantly improves upon current benchmark studies that consider model accuracy and computational costs only.  Furthermore, to our knowledge, this is the first ML benchmark analysis that evaluates algorithms for predicting rare events in chemical process safety.  For the weighting coefficients considered in Eq. (\ref{eq_13}), the gradient-boosting frameworks; i.e., XGBoost, LightGBM, and CatBoost, outperform other algorithms, achieving strong predictive performance at low computational costs, also providing relatively favorable metrics for alarm-system efficiency.  DNN and TabNet require more computational resources that are not justified by their model accuracy, although DNN offers fast deployment across all datasets.  Despite much promise for tabular datasets, TabNet consistently performs poorly across all datasets.  Additionally, Linear SVR and kNN compensate for lower model accuracies by having low computational costs, but, along with RF, are outperformed consistently by the gradient-boosting frameworks when all metrics in the $Cost$ are considered.  Moreover, based on the global rankings recorded in Figure \ref{fig_9} that consider 500 randomly-sampled combinations of weighting coefficients, CatBoost is the most-optimal algorithm across all datasets and evaluation metrics, followed by XGBoost, LightGBM, and DNN.  Note that increased \textit{RMSE}, $\Delta p$ and \textit{Total Alarms} may contribute potentially to increased false alarm and missed alarm rates.  Additionally, higher model deployment times may result in a lag between real-time process variable measurements and $p_{\text{B}}$ predictions.  Hence, such comprehensive benchmark frameworks will aid the operator in selecting the most-optimal ML algorithm for process monitoring and predictive maintenance against rare abnormal events, improving their effectiveness in ensuring safety and reliability.

\paragraph{} Note that despite these encouraging findings, a few limitations should be addressed.  For instance, herein, all datasets generated using BG-FFS are based on first-principles models (i.e., material and energy-balance ODEs, reaction kinetics, and the like), with assumptions simplifying the process models \cite{sudarshan_understanding_2021, sudarshan_path-sampling_2024}.  Additionally, the analyses presented herein utilize a specific 2-level alarm system (see Table \ref{table_2}).  For more-comprehensive analyses, it is important to consider several alarm-system combinations, but this would require significant computational costs.  In future research, hybrid computational models (e.g., physics-informed neural networks; i.e., PINNs) involving underlying physics, coupled with plant data from sensors, alarm databases, and the like, should be developed.  Additionally, $\Delta p$ (i.e., see Eq. (\ref{eq_10})) should be utilized to develop more-intelligent, automated/semi-automated, alarm rationalization strategies, a significant improvement compared to the framework developed in our prior research \cite{sudarshan_alarm_2024}.

\section*{Acknowledgments}
The NSF CBET funding for our grant, 2220276, is greatly appreciated.  Thanks are extended to Amish Patel, Ulku Oktem, and Jeff Arbogast, who provided advice throughout this research.

\appendix
\renewcommand\thesection{A}  
\section*{Appendix}  

\renewcommand\thesubsection{\thesection.\arabic{subsection}}  

\setcounter{table}{0}  
\renewcommand{\thetable}{\thesection.\arabic{table}}  

\setcounter{figure}{0}
\renewcommand{\thefigure}{\thesection.\arabic{figure}}  

\begin{figure}[h]
    \centering
    \begin{subfigure}{0.42\textwidth}  
        \centering
        \includegraphics[width=\linewidth]{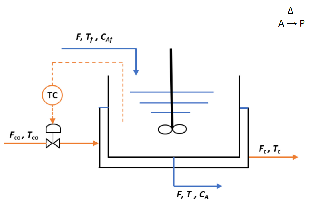}  
        \caption{}
        \label{fig_a1a}
    \end{subfigure}
    \hfill
    \begin{subfigure}{0.42\textwidth}  
        \centering
        \includegraphics[width=\linewidth]{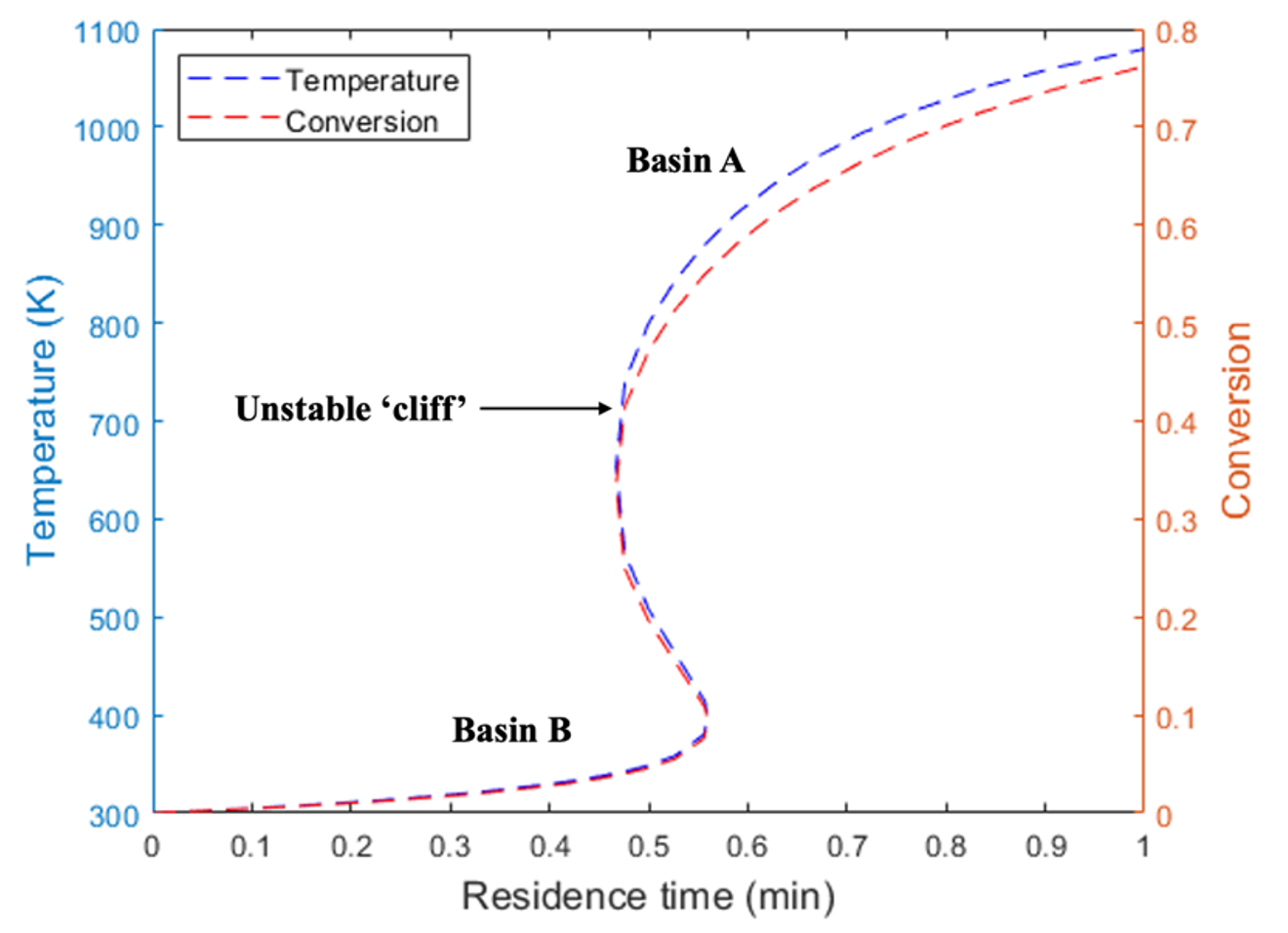}  
        \caption{}
        \label{fig_a1b}
    \end{subfigure}
    \vspace{1em}  
    \begin{subfigure}{0.42\textwidth}  
        \centering
        \includegraphics[width=\linewidth]{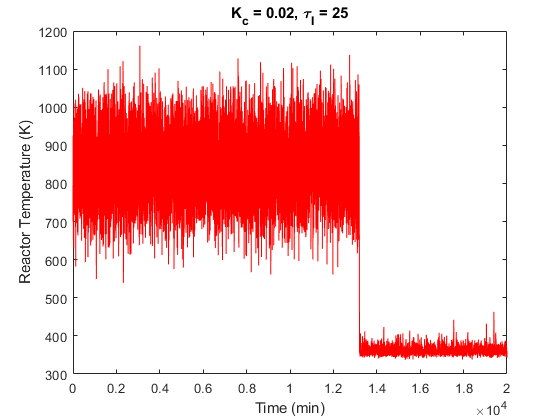}  
        \caption{}
        \label{fig_a1c}
    \end{subfigure}
    \caption{a) Schematic of the PI-controlled exothermic CSTR; (b) Process under steady-state operation; (c) Process showing the “noise”-induced un-postulated abnormal transition.}
    \label{fig_a1}
\end{figure}

\subsection{Proportional-Integral (PI)-Controlled Exothermic CSTR}
\label{sec_A.1}
 Previously, novel, multivariate alarm systems were developed using path-sampling and predictive modeling for a relatively simple P-only controlled exothermic CSTR, followed by the alarm rationalization-DRAn integrated framework for further enhancement \cite{sudarshan_multivariate_2023, sudarshan_alarm_2024}.  In this Appendix \ref{sec_A.1}, we provide a brief summary of the process model.  Figure \ref{fig_a1a} shows a schematic of the model for a Proportional-Integral (PI)-controlled exothermic CSTR, with first-order kinetics, i.e., A $\rightarrow$ P.  The assumptions for this ideal process model include: i) Constant residence time; ii) Incompressible flow; iii) Complete back-mixing.  The model attempts to control the reactor temperature, $T$, by manipulating the coolant flow-rate, $F_\text{C}$.  Additionally, Figure \ref{fig_a1b} shows the steady-state behavior for the model, with multiple steady-states observed for residence time; i.e., $\tau$ $\epsilon$ [0.47, 0.56] min. Two stable steady-states are observed; i.e., at the high conversion-high temperature basin ‘A’, and low-conversion, low-temperature basin ‘B’.  In between these two, a wide unstable “cliff” exists, such that if the process operates anywhere near this cliff, with sufficient input perturbation, the process shifts rapidly to either of the two stable regions\cite{moskowitz_understanding_2018, sudarshan_understanding_2021}. 
 \paragraph{} The governing equations for the PI-controlled process are:

\begin{equation}
    \begin{aligned}
    V \frac{dC_{\text{A}}}{dt} = \frac{V}{\tau} \left(C_{\text{Af}} - C_{\text{A}} + \eta \right) - Vk_0 \exp\left(\frac{-E}{RT}\right) C_{\text{A}}
    \end{aligned}
    \tag{A.1}
    \label{eq_a1}
\end{equation}

\begin{equation}
\begin{aligned}
    \rho V c_\text{p} \frac{dT}{dt} &= \frac{\rho V c_\text{p}}{\tau} \left(T_\text{f} - T\right) - V\Delta H k_0 \exp\left(\frac{-E}{RT}\right) C_{\text{A}} \\
    &\, + UA(T_\text{C} - T)
\end{aligned}
\tag{A.2}
\label{eq_a2}
\end{equation}

\begin{equation}
    \frac{dT_{\text{C}}}{dt} = \frac{F_{\text{C}}}{V_\text{j}} (T_{\text{C0}} - T_{\text{C}}) - \frac{UA}{\rho_\text{w} V_\text{j} c_{\text{pw}}} (T_{\text{C}} - T)
    \tag{A.3}
    \label{eq_a3}
\end{equation}

\begin{equation}
    F_{\text{C}} = F_{\text{C0}} + K_{\text{C}} (T - T_{\text{SP}} - \frac{e_{\text{I}}}{\tau_{\text{I}}}); \, 30 \leq F_{\text{C}} \leq 70
    \tag{A.4}
    \label{eq_a4}
\end{equation}

\begin{equation}
    \frac{de_{\text{I}}}{dt} = T_{\text{SP}} - T
    \tag{A.5}
    \label{eq_a5}
\end{equation}

\begin{equation}
    C_{\text{A0}} = 1.2 \text{ kmol/m}^3; \quad T_0 = 700 \text{ K}
    \tag{A.6}
    \label{eq_a6}
\end{equation}

where $C_\text{A}$ is the concentration of reactant A; $T$ is the reactor temperature; $C_\text{Af}$ and $T_\text{f}$ are the concentration and temperature for the reactant feed stream; $F_\text{C0}$ is the cooling water flow-rate at steady state, $K_\text{C}$ is the controller gain, $e_\text{I}$ is the integral error and $\tau_\text{I}$ is the integral time constant; $T_\text{C0}$ is the inlet temperature of the cooling water, $C_\text{A0}$ is the initial value for the concentration, $T_0$ is the initial value for the temperature, $T_\text{SP}$ is the set-point temperature for the controller, $V_\text{reactor}$ is the volume of the reactor, $U$ is the overall heat-transfer coefficient, $A$ is the heat-transfer area, $\Delta H$ is the heat of reaction, $\rho$ is the feed density, $c_\text{p}$ is the heat capacity of the feed stream, $V_\text{j}$ is the volume of the cooling-water jacket, $\rho_\text{w}$ is the density of the cooling water, and $c_\text{pw}$ is the specific-heat capacity of the cooling water (refer to Table \ref{table_A1}). 
\paragraph{}To induce un-postulated abnormal transitions from the desirable basin “A” to the undesirable and unreliable basin “B,” statistical “noise”-induced perturbations; i.e., $\eta$, are utilized. Note that $\eta$ is sampled randomly at every integration time-step from a normal distribution; i.e., $\eta \sim N(\mu,\sigma_\eta^2)$, with a mean; i.e., $\mu = 0$, and variance; i.e., $\sigma_\eta^2 = 0.02$. Figure \ref{fig_a1c} shows a dynamic simulation of the process under noisy operation, showing the un-postulated abnormal transition from basins A to B – several such trajectories are simulated efficiently using BGFFS, followed by calculations of the committer probability; i.e., $p_\text{B}$, defined as the probability that a trajectory fired from a given location reaches or “commits” to the undesirable region.

\begin{table}[h]
\centering
\caption{Process Constants and Parameters for PI-Controlled Exothermic CSTR}
\vspace{1em}  
\renewcommand{\arraystretch}{1.5}  
\begin{tabular}{l c c}  
\hline
\textbf{Parameter} & \textbf{Value} & \textbf{Unit} \\
\hline
$A$ & 30 & m$^2$ \\
$C_{\text{Af}}$ & 2 & kmol/m$^3$ \\
$c_{\text{p}} = c_{\text{pw}}$ & 4 & kJ/(kg$\cdot$K) \\
$E$ & $1.50 \times 10^4$ & kJ/kmol \\
$F_{\text{C0}}$ & 50 & m$^3$/min \\
$k_0$ & 17.038 & 1/min \\
$R$ & 8.314 & kJ/(kmol$\cdot$K) \\
$T_{\text{C0}}$ & 300 & K \\
$T_{\text{f}}$ & 300 & K \\
$T_{\text{SP}}$ & 800 & K \\
$U$ & 100 & kJ/(min$\cdot$K$\cdot$m$^2$) \\
$V_{\text{reactor}}$ $= V_{\text{j}}$ & 10 & m$^3$ \\
$\Delta H$ & $-2.20 \times 10^6$ & kJ/kmol \\
$\rho = \rho_{\text{w}}$ & 1,000 & kg/m$^3$ \\
$K_{\text{C}}$ & 0.02 & m$^3$/min$\cdot$K \\
$\tau_{\text{I}}$ & 25 & min \\
\hline
\end{tabular}
\label{table_A1}
\end{table}


\begin{figure*}[h]
    \centering
    \begin{subfigure}{0.47\textwidth}  
        \centering
        \includegraphics[width=\linewidth]{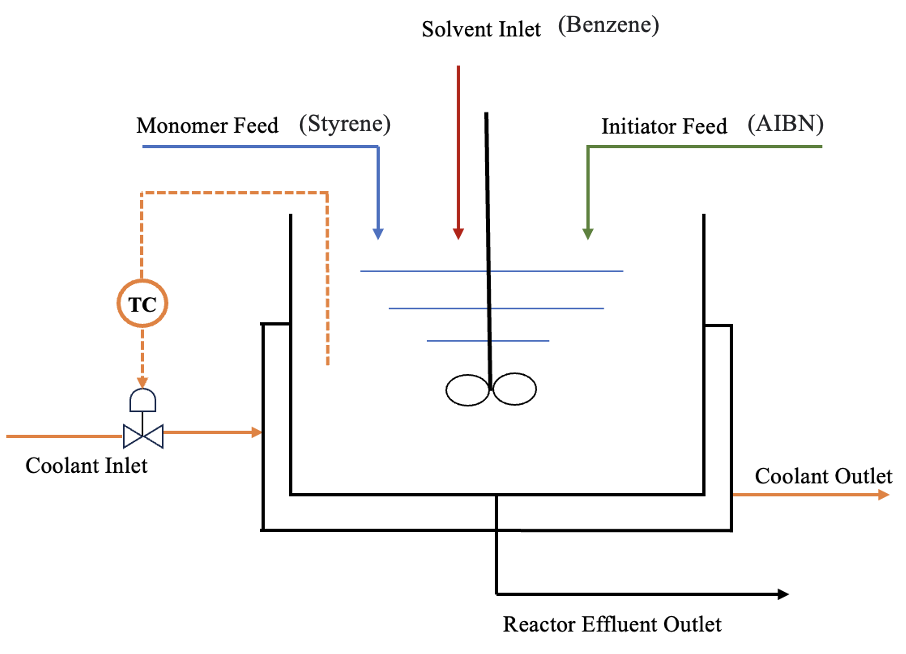}  
        \caption{}
        \label{fig_a2a}
    \end{subfigure}
    \hfill
    \begin{subfigure}{0.47\textwidth}  
        \centering
        \includegraphics[width=\linewidth]{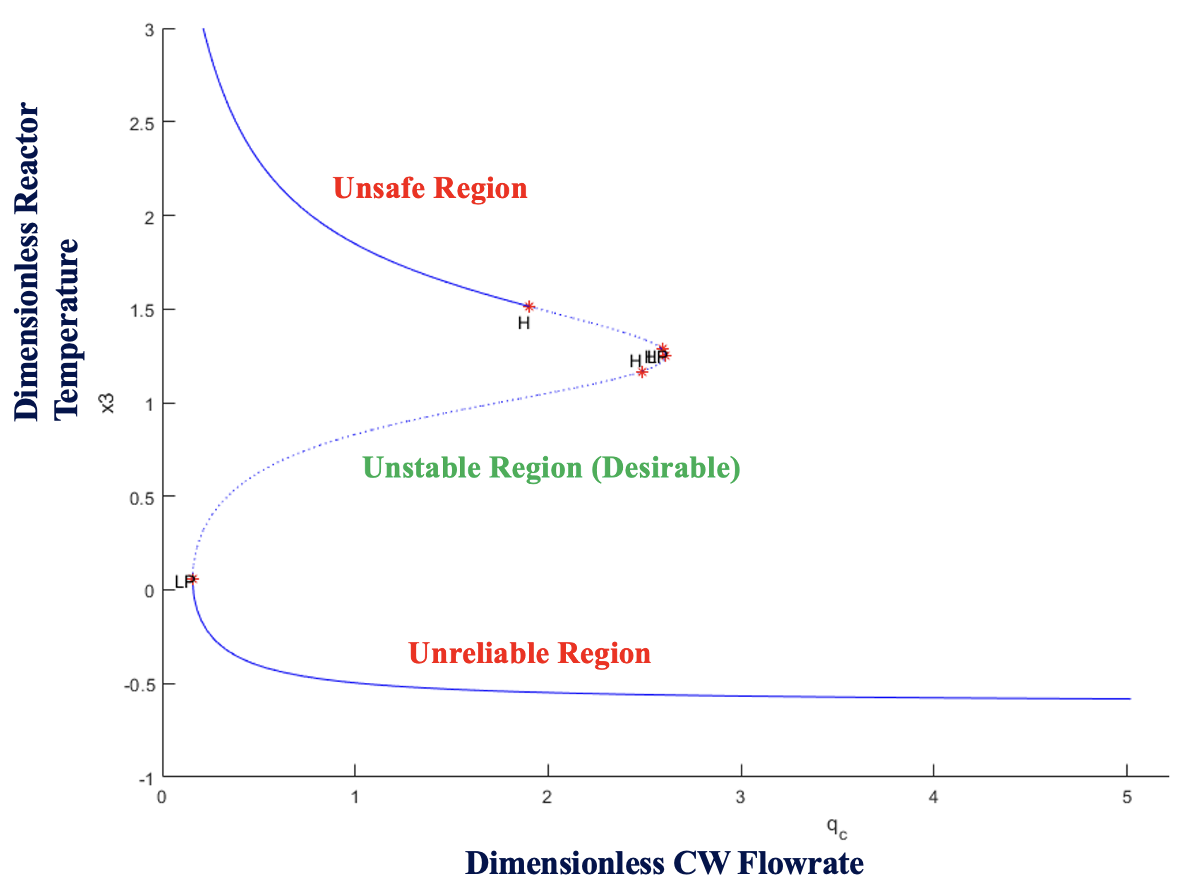}  
        \caption{}
        \label{fig_a2b}
    \end{subfigure}
    \begin{subfigure}{0.47\textwidth}  
        \centering
        \includegraphics[width=\linewidth]{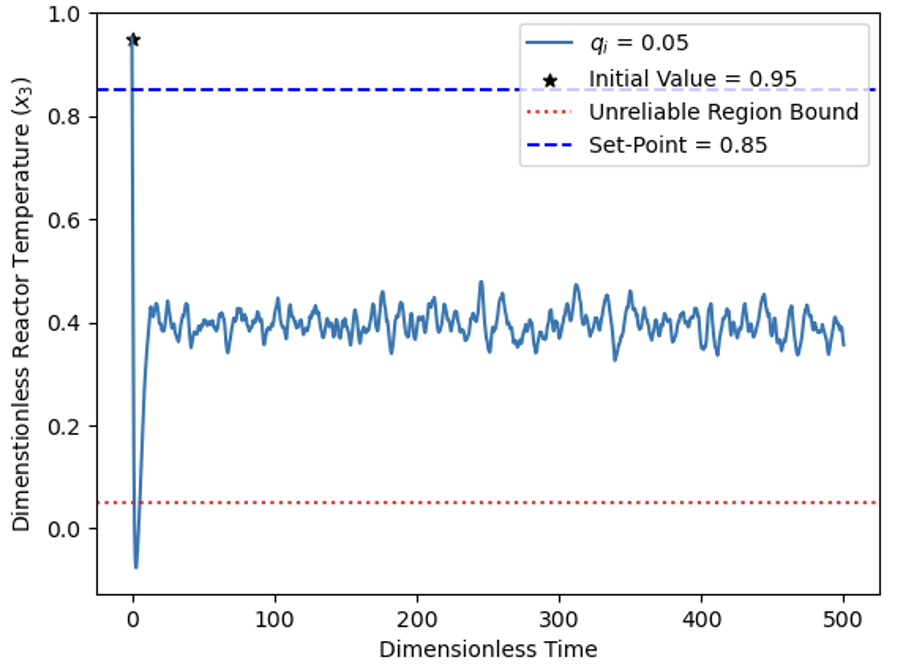}  
        \caption{}
        \label{fig_a2c}
    \end{subfigure}
    \hfill
    \begin{subfigure}{0.47\textwidth}  
        \centering
        \includegraphics[width=\linewidth]{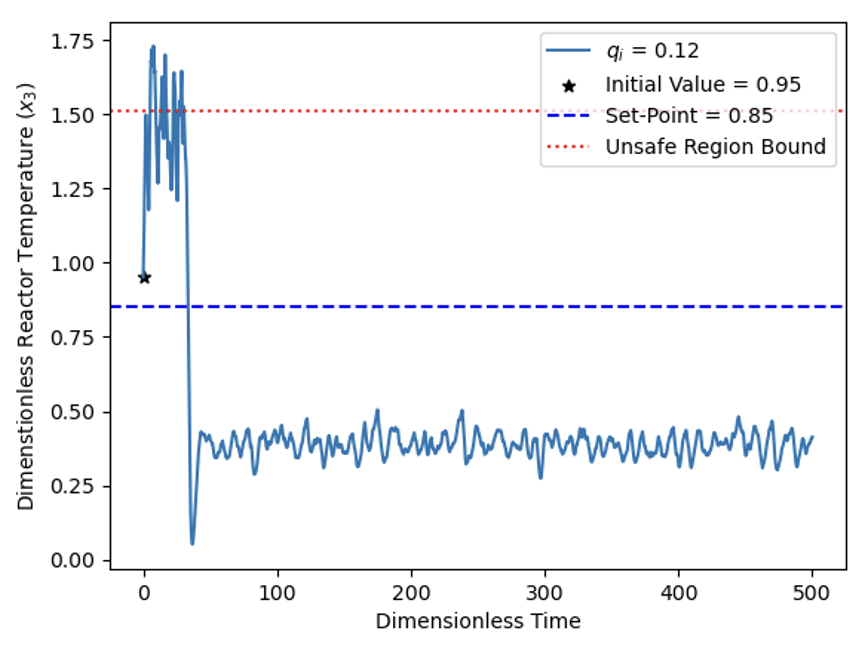}  
        \caption{}
        \label{fig_a2d}
    \end{subfigure}
    \caption{(a) Schematic of PID-controlled polystyrene CSTR; (b) Process under steady-state operation, showing the desirable unstable and undesirable stable region. Note the key bifurcation points; i.e., limit points, LP, and Hopf bifurcation points, H; (c) Process under dynamic operation, showing the un-postulated transition towards the unreliable region; (d) Process under dynamic operation, showing the un-postulated transition towards the unsafe region.}
    \label{fig_a2}
\end{figure*}

\subsection{Proportional-Integral-Derivative (PID)-Controlled Polymerization CSTR}
\label{sec_A.2}
Given the limitations of the alarm systems developed for the P-only controlled exothermic CSTR, in recent research, improved, dynamic, bidirectional multivariate alarm systems were developed for a PID-controlled polystyrene CSTR, capable of handling un-postulated abnormal shifts to multiple undesirable regions; i.e., unsafe and unreliable \cite{sudarshan_path-sampling_2024}.  In this Appendix \ref{sec_A.2}, we provide a brief summary of the process model.  Figure \ref{fig_a2a} shows a schematic of the model.  The governing equations for the dimensionless PID-controlled polystyrene CSTR are:

\begin{equation}
\frac{dx_1}{d\tau} = q_\text{i} x_\text{1f} - (q_\text{i} + q_\text{m} + q_\text{s}) x_1 - \phi_\text{d} \kappa_\text{d}(x_3) x_1
\tag{A.7}
\label{eq_a7}
\end{equation}

\begin{align}
\frac{dx_2}{d\tau} &= q_\text{m} (x_\text{2f} + \eta) - (q_\text{i} + q_\text{m} + q_\text{s}) x_2 \nonumber \\
&\quad - \phi_\text{p} \kappa_\text{p}(x_3) x_2 x_5; \quad \eta \sim N(0, \sigma_\eta^2)
\tag{A.8}
\label{eq_a8}
\end{align}

\begin{align}
\frac{dx_3}{d\tau} &= (q_\text{i} + q_\text{m} + q_\text{s}) (x_\text{3f} - x_3) \nonumber \\
&\quad + \beta \phi_\text{p} \kappa_\text{p}(x_3) x_2 x_5 \nonumber \\
&\quad - \delta(x_3 - x_4)
\tag{A.9}
\label{eq_a9}
\end{align}

\begin{equation}
\frac{dx_4}{d\tau} = \delta_\text{1} \left[ q_\text{c} (x_\text{4f} - x_4) + \delta \delta_\text{2} (x_3 - x_4) \right]
\tag{A.10}
\label{eq_a10}
\end{equation}

\begin{equation}
x_5 = \sqrt{\frac{2f \phi_\text{d} \kappa_\text{d}(x_3) x_1}{\phi_\text{t} \kappa_\text{t}(x_3)}}
\tag{A.11}
\label{eq_a11}
\end{equation}

\begin{align}
q_\text{c} &= q_\text{c,0} - K_\text{c} \left[ (x_\text{3,sp} - x_3) 
+ \frac{1}{\tau_\text{I}} \int_0^t (x_\text{3,sp} - x_3) dt' \right. \nonumber \\
&\quad \left. + \tau_\text{D} \frac{d(x_\text{3,sp} - x_3)}{dt} \right]
\tag{A.12}
\label{eq_a12}
\end{align}

\begin{equation}
0 \leq q_\text{c} \leq 5
\tag{A.13}
\label{eq_a13}
\end{equation}

\begin{equation}
\begin{split}
x_\text{1,0} = 0.0041; \, x_\text{2,0} = 0.2156; \\
x_\text{3,0} = 0.951; \, x_\text{4,0} = -1.1191; \, q_\text{c,0} = 1.5  
\end{split}
\tag{A.14}
\label{eq_a14}
\end{equation}

\begin{equation}
\kappa_\text{d}(x_3) = \exp\left(\frac{\gamma_\text{d} x_3}{1 + \frac{x_3}{\gamma_\text{p}}}\right)
\tag{A.15}
\label{eq_a15}
\end{equation}

\begin{equation}
\kappa_\text{t}(x_3) = \exp\left(\frac{\gamma_\text{t} x_3}{1 + \frac{x_3}{\gamma_\text{p}}}\right)
\tag{A.16}
\label{eq_a16}
\end{equation}

\begin{equation}
\kappa_\text{p}(x_3) = \exp\left(\frac{x_3}{1 + \frac{x_3}{\gamma_\text{p}}}\right)
\tag{A.17}
\label{eq_a17}
\end{equation}

\begin{figure*}[h]
    \centering
    \includegraphics[width=0.75\linewidth]{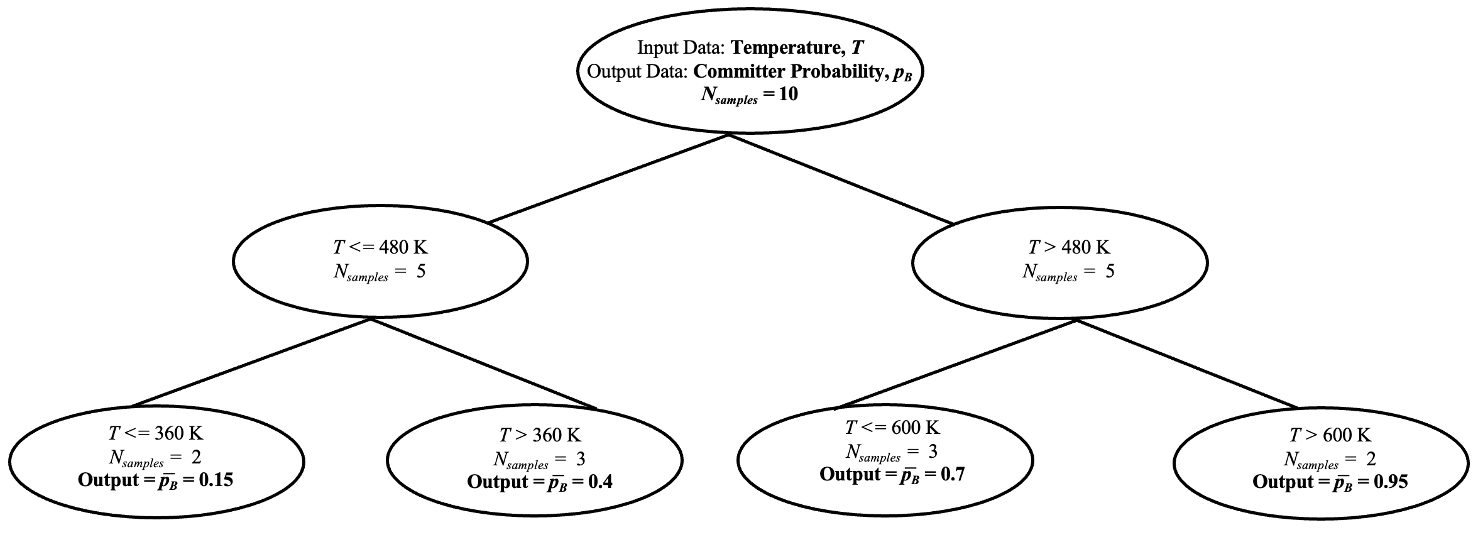}
    \caption{Schematic of a decision tree, a flowchart-like model that helps make decisions by answering a series of questions based on the input variables (e.g., temperature) of the data, ultimately leading to a decision or prediction.}
    \label{fig_a3}
\end{figure*}

where $x_\text{1}$, $x_\text{2}$, $x_\text{3}$, $x_\text{4}$, $x_\text{5}$ are the dimensionless initiator concentration, monomer concentration, reactor temperature, coolant temperature, and concentration of growing polymer; $q_\text{i}$, $q_\text{m}$, $q_\text{s}$, $q_\text{c}$ are the dimensionless flow rates for initiator, monomer, solvent, and coolant streams; $\phi_\text{d}$, $\phi_\text{p}$, $\phi_\text{t}$ are Damköhler numbers for initiator decomposition, propagation, and termination; $\gamma_\text{d}$, $\gamma_\text{p}$, $\gamma_\text{t}$ are dimensionless activation energies for initiator decomposition, propagation, and termination; $\beta$ is the dimensionless heat of reaction, $\delta$ is the dimensionless heat-transfer coefficient, $\delta_\text{1}$ is the dimensionless reactor volume, $\delta_\text{2}$ is the dimensionless specific heat, $f$ is the initiator efficiency; $x_\text{1f}$, $x_\text{2f}$, $x_\text{3f}$, $x_\text{4f}$ are the dimensionless initiator feed concentration, monomer feed concentration, reactor feed temperature, and coolant feed temperature; $x_\text{1,0}$, $x_\text{2,0}$, $x_\text{3,0}$, $x_\text{4,0}$, $q_\text{c,0}$ represent initial values. Similar to the PI-controlled exothermic CSTR, statistical “noise,” i.e., $\eta$, is sampled at every integration time-step from a normal distribution, $N(\mu, \sigma_\eta^2)$, with a mean $\mu = 0$, and variance $\sigma_\eta^2 = 0.0014$. Note that $\eta$ is added to the dimensionless monomer concentration, $x_\text{2f}$, in Eq.(\ref{eq_a8}). Figure \ref{fig_a2b} shows the process under steady-state operation, with the process constants and parameters in Table \ref{table_A2}; the intermediate, unstable region is the desirable region, with possible un-postulated abnormal shifts to both unsafe and unreliable regions, which are stable. Figure \ref{fig_a2} c-d shows the process under noisy operation, with un-postulated abnormal transitions observed towards both undesirable regions. In Figure \ref{fig_a2c}, the process moves from $x_\text{3} = 0.95$ in the unstable region to a brief visit to the unreliable region before settling in the unstable region. In Figure \ref{fig_a2d}, it moves first to the unsafe region before returning to the unstable region. For more details regarding the process model and these noisy trajectories, please refer to \cite{russo_operability_1998} and \cite{sudarshan_path-sampling_2024}.

\begin{table}[h]
\centering
\caption{Process Constants and Parameters for PID-controlled Polystyrene CSTR}
\vspace{1em}  
\renewcommand{\arraystretch}{1.5}  
\begin{tabular}{l c}  
\hline
\textbf{Parameter} & \textbf{Value} \\
\hline
$q_{\text{i}}$ & 0.1 \\
$q_{\text{m}}$ & 0.4 \\
$q_{\text{s}}$ & 0.48571 \\
$\phi_{\text{d}}$ & 0.01688 \\
$\phi_{\text{p}}$ & $2.1956 \times 10^7$ \\
$\phi_{\text{t}}$ & $9.6583 \times 10^{12}$ \\
$x_{\text{1f}}$ & 0.06769 \\
$x_{\text{2f}}$ & 1.0 \\
$x_{\text{3f}}$ & 0.0 \\
$x_{\text{4f}}$ & -1.5 \\
$\delta$ & 0.74074 \\
$\delta_1$ & 0.90569 \\
$\delta_2$ & 0.37256 \\
$\beta$ & 13.17936 \\
$f$ & 0.6 \\
$x_{3,\text{sp}}$ & 0.85 \\
$K_{\text{c}}$ & 50 \\
$\tau_{\text{D}}$ & 0.9 \\
$\tau_{\text{I}}$ & 5 \\
\hline
\end{tabular}
\label{table_A2}
\end{table}

\subsection{Computational Specifications}
\label{sec_A.3}
Note that all analyses and results presented herein were conducted on a Desktop computer, having specifications:

i)	Operating System (OS): Linux via WSL (i.e., Windows Subsystem for Linux) 2. \par
ii)	CPU: $12^{\text{th}}$ - generation Intel i7-12700K with 12 cores (8 performance + 4 efficiency), 32 GB DDR5 RAM. \par
iii)	GPU: NVIDIA RTX 3060 Ti, 8 GB RAM.\par

\paragraph{}The Python programming language (version 3.9) is utilized, leveraging several powerful open-source software packages, including: NumPy \cite{harris_array_2020}, Pandas \cite{mckinney_data_2010}, SciPy \cite{virtanen_scipy_2020}, Matplotlib \cite{hunter_matplotlib_2007}, Scikit-Learn \cite{pedregosa_scikit-learn_2011}, XGBoost \cite{chen_xgboost_2016}, LightGBM \cite{ke_lightgbm_2017}, CatBoost \cite{prokhorenkova_catboost_2018}, Numba \cite{lam_numba_2015}, Optuna \cite{akiba_optuna_2019}, to name a few.  For developing DNNs, PyTorch \cite{paszke_pytorch_2019} is utilized; with PyTorch-TabNet \cite{arik_tabnet_2020, dreamquark_pytorch_tabnet_2019} utilized to develop TabNet models.  Also, for GPU-acceleration during model development, the NVIDIA Compute Unified Device Architecture (CUDA) toolkit \cite{nvidia_nvidia_2022} is utilized.  Note that while XGBoost, LightGBM, CatBoost, PyTorch and PyTorch-TabNet have native support for CUDA-based GPU-acceleration, the Scikit-Learn-based implementations of Linear SVR, kNN, and RF does not support GPU acceleration natively – to address this, \textit{RAPIDS AI}; i.e., an open-source suite of software packages developed for GPU acceleration \cite{rapids_development_team_rapids_2023}, is utilized, with cuML and cuDF packages providing GPU acceleration for Linear SVR, kNNs, and RF.

\begin{table}[h]
\centering
\caption{Example Dataset for the Decision Tree Trained in Figure \ref{fig_a3}}
\vspace{1em}
\renewcommand{\arraystretch}{1.5}
\begin{tabular}{cc}
\hline
\textbf{Temperature, $T$ (K)} & \textbf{Committer Probability, $p_\text{B}$} \\
\hline
300 & 0.1 \\
340 & 0.2 \\
380 & 0.3 \\
420 & 0.4 \\
460 & 0.5 \\
500 & 0.6 \\
540 & 0.7 \\
580 & 0.8 \\
620 & 0.9 \\
660 & 1.0 \\
\hline
\end{tabular}
\label{table_A3}
\end{table}

\subsection{Simple Example of a Decision Tree}
\label{sec_A.4}
Figure \ref{fig_a3} shows a schematic for a decision tree involving an example dataset created for demonstration purposes only, containing just 10 samples ($N_\text{samples} = 10$), to predict the committer probability; i.e., $p_\text{B}$ as a function of temperature, $T$. For this dataset, Table \ref{table_A3} shows $p_\text{B}$ and $T$ for each of the samples. At each iteration, the optimum split threshold for $T$ is computed using the variance reduction method described in \cite{breiman_chapter_2017}. On this basis, the data is split further, with the splitting process terminated when insufficient data remain, after which the average $p_\text{B}$ is returned. For instance, at the first split, the optimum split threshold for $T$ is computed as $T = 480 \, \text{K}$ – then, the data are divided into two sets; i.e., 5 samples for which $T \leq 480 \, \text{K}$, and 5 samples for which $T > 480 \, \text{K}$. The splitting process continues for both sets until the number of samples remaining in a set is $\leq N_\text{min}$ ($N_\text{min}$ is the threshold for the number of samples in a set to stop splitting; e.g., in Figure \ref{fig_a3}, $N_\text{min} = 3$), thereby returning the average $p_\text{B}$ for that set. To check for consistency with the data in Table \ref{table_A3}, note the average $p_\text{B}$ values returned at the end of the decision tree in Figure \ref{fig_a3} – for instance, there are 2 samples for which $T \leq 360$ (i.e., $T = \{300, 340\}$; $p_\text{B} = \{0.1, 0.2\}$), with the average $p_\text{B}$ for these samples = 0.15, consistent with Figure \ref{fig_a3}. Similarly, there are 3 samples for which $360 < T \leq 480$ (i.e., $T = \{380, 420, 460\}$; $p_\text{B} = \{0.3, 0.4, 0.5\}$), with their average $p_\text{B} = 0.4$.

\bibliographystyle{ieeetr}
\bibliography{arxiv_benchmark_updated}

\begin{thebibliography}{100}

\bibitem{crafts_explaining_2011}
N.~Crafts, ``Explaining the first {Industrial} {Revolution}: two views,'' {\em European Review of Economic History}, vol.~15, no.~1, pp.~153--168, 2011.

\bibitem{mohajan_first_2019}
H.~Mohajan, ``The {First} {Industrial} {Revolution}: {Creation} of a {New} {Global} {Human} {Era},'' {\em Journal of Social Sciences and Humanities}, vol.~5, no.~4, pp.~377--387, 2019.

\bibitem{mokyr_second_1998}
J.~Mokyr and R.~H. Strotz, ``The second industrial revolution, 1870-1914,'' {\em Storia dell’economia Mondiale}, vol.~21945, no.~1, 1998.

\bibitem{mohajan_third_2021}
H.~Mohajan, ``Third {Industrial} {Revolution} {Brings} {Global} {Development},'' {\em Journal of Social Sciences and Humanities}, vol.~7, no.~4, pp.~239--251, 2021.

\bibitem{naboni_third_2015}
R.~Naboni and I.~Paoletti, ``The {Third} {Industrial} {Revolution},'' in {\em Advanced {Customization} in {Architectural} {Design} and {Construction}}, pp.~7--27, Cham: Springer International Publishing, 2015.

\bibitem{belli_toward_2019}
L.~Belli, L.~Davoli, A.~Medioli, P.~L. Marchini, and G.~Ferrari, ``Toward {Industry} 4.0 {With} {IoT}: {Optimizing} {Business} {Processes} in an {Evolving} {Manufacturing} {Factory},'' {\em Frontiers in ICT}, vol.~6, 2019.

\bibitem{soori_internet_2023}
M.~Soori, B.~Arezoo, and R.~Dastres, ``Internet of things for smart factories in industry 4.0, a review,'' {\em Internet of Things and Cyber-Physical Systems}, vol.~3, pp.~192--204, 2023.

\bibitem{becue_artificial_2021}
A.~Bécue, I.~Praça, and J.~Gama, ``Artificial intelligence, cyber-threats and {Industry} 4.0: challenges and opportunities,'' {\em Artificial Intelligence Review}, vol.~54, no.~5, pp.~3849--3886, 2021.

\bibitem{candanedo_machine_2018}
I.~S. Candanedo, E.~H. Nieves, S.~R. González, M.~T.~S. Martín, and A.~G. Briones, ``Machine {Learning} {Predictive} {Model} for {Industry} 4.0,'' in {\em Knowledge {Management} in {Organizations}}, Communications in {Computer} and {Information} {Science}, (Cham), pp.~501--510, Springer International Publishing, 2018.

\bibitem{dingli_artificial_2021}
A.~Dingli, F.~Haddod, and C.~Kl{\"u}ver, {\em Artificial intelligence in industry 4.0}, vol.~928.
\newblock Springer, 2021.

\bibitem{culot_addressing_2019}
G.~Culot, F.~Fattori, M.~Podrecca, and M.~Sartor, ``Addressing {Industry} 4.0 {Cybersecurity} {Challenges},'' {\em IEEE Engineering Management Review}, vol.~47, no.~3, pp.~79--86, 2019.

\bibitem{ervural_overview_2018}
B.~C. Ervural and B.~Ervural, ``Overview of {Cyber} {Security} in the {Industry} 4.0 {Era},'' in {\em Industry 4.0: {Managing} {The} {Digital} {Transformation}}, Springer {Series} in {Advanced} {Manufacturing}, pp.~267--284, Cham: Springer International Publishing, 2018.

\bibitem{hashimoto_safety_2013}
Y.~Hashimoto, T.~Toyoshima, S.~Yogo, M.~Koike, T.~Hamaguchi, S.~Jing, and I.~Koshijima, ``Safety securing approach against cyber-attacks for process control system,'' {\em Computers \& Chemical Engineering}, vol.~57, pp.~181--186, 2013.

\bibitem{gokalp_big_2016}
M.~O. Gokalp, K.~Kayabay, M.~A. Akyol, P.~E. Eren, and A.~Koçyiğit, ``Big {Data} for {Industry} 4.0: {A} {Conceptual} {Framework},'' in {\em 2016 {International} {Conference} on {Computational} {Science} and {Computational} {Intelligence} ({CSCI})}, pp.~431--434, 2016.

\bibitem{kim_review_2017}
J.~H. Kim, ``A {Review} of {Cyber}-{Physical} {System} {Research} {Relevant} to the {Emerging} {IT} {Trends}: {Industry} 4.0, {IoT}, {Big} {Data}, and {Cloud} {Computing},'' {\em Journal of Industrial Integration and Management}, vol.~02, no.~03, p.~1750011, 2017.

\bibitem{barata_industry_2023}
J.~Barata and I.~Kayser, ``Industry 5.0 – {Past}, {Present}, and {Near} {Future},'' {\em Procedia Computer Science}, vol.~219, pp.~778--788, 2023.

\bibitem{demir_industry_2019}
K.~A. Demir, G.~Döven, and B.~Sezen, ``Industry 5.0 and {Human}-{Robot} {Co}-working,'' {\em Procedia Computer Science}, vol.~158, pp.~688--695, 2019.

\bibitem{ghobakhloo_behind_2023}
M.~Ghobakhloo, M.~Iranmanesh, M.-L. Tseng, A.~Grybauskas, A.~Stefanini, and A.~Amran, ``Behind the definition of {Industry} 5.0: a systematic review of technologies, principles, components, and values,'' {\em Journal of Industrial and Production Engineering}, vol.~40, no.~6, pp.~432--447, 2023.

\bibitem{raja_santhi_industry_2023}
A.~Raja~Santhi and P.~Muthuswamy, ``Industry 5.0 or industry 4.{0S}? {Introduction} to industry 4.0 and a peek into the prospective industry 5.0 technologies,'' {\em International Journal on Interactive Design and Manufacturing (IJIDeM)}, vol.~17, no.~2, pp.~947--979, 2023.

\bibitem{hailwood_learning_2016}
M.~Hailwood, ``Learning from {Accidents} – {Reporting} is not {Enough},'' {\em Chemical Engineering Transactions}, vol.~48, pp.~709--714, 2016.

\bibitem{broughton_bhopal_2005}
E.~Broughton, ``The {Bhopal} disaster and its aftermath: a review,'' {\em Environmental Health}, vol.~4, no.~1, p.~6, 2005.

\bibitem{gupta_bhopal_2002}
J.~P. Gupta, ``The {Bhopal} gas tragedy: could it have happened in a developed country?,'' {\em Journal of Loss Prevention in the Process Industries}, vol.~15, no.~1, pp.~1--4, 2002.

\bibitem{sriramachari_bhopal_2004}
S.~Sriramachari, ``The {Bhopal} gas tragedy: {An} environmental disaster,'' {\em Current Science}, vol.~86, no.~7, pp.~905--920, 2004.

\bibitem{saenko_chernobyl_2011}
V.~Saenko, V.~Ivanov, A.~Tsyb, T.~Bogdanova, M.~Tronko, Y.~Demidchik, and S.~Yamashita, ``The {Chernobyl} {Accident} and its {Consequences},'' {\em Clinical Oncology}, vol.~23, no.~4, pp.~234--243, 2011.

\bibitem{holmstrom_csb_2006}
D.~Holmstrom, F.~Altamirano, J.~Banks, G.~Joseph, M.~Kaszniak, C.~Mackenzie, R.~Shroff, H.~Cohen, and S.~Wallace, ``{CSB} investigation of the explosions and fire at the {BP} texas city refinery on {March} 23, 2005,'' {\em Process Safety Progress}, vol.~25, no.~4, pp.~345--349, 2006.

\bibitem{beyer_environmental_2016}
J.~Beyer, H.~C. Trannum, T.~Bakke, P.~V. Hodson, and T.~K. Collier, ``Environmental effects of the {Deepwater} {Horizon} oil spill: {A} review,'' {\em Marine Pollution Bulletin}, vol.~110, no.~1, pp.~28--51, 2016.

\bibitem{labib_learning_2015}
A.~Labib and M.~J. Harris, ``Learning how to learn from failures: {The} {Fukushima} nuclear disaster,'' {\em Engineering Failure Analysis}, vol.~47, pp.~117--128, 2015.

\bibitem{us_chemical_safety_and_hazard_investigation_board_bp_2007}
U.S.C.S.B, ``{BP} {America} ({Texas} {City}) {Refinery} {Explosion},'' tech. rep., United States Chemical Safety and Hazard Investigation Board, 2007.

\bibitem{lavecchia_machine-learning_2015}
A.~Lavecchia, ``Machine-learning approaches in drug discovery: methods and applications,'' {\em Drug Discovery Today}, vol.~20, no.~3, pp.~318--331, 2015.

\bibitem{vamathevan_applications_2019}
J.~Vamathevan, D.~Clark, P.~Czodrowski, I.~Dunham, E.~Ferran, G.~Lee, B.~Li, A.~Madabhushi, P.~Shah, M.~Spitzer, and S.~Zhao, ``Applications of machine learning in drug discovery and development,'' {\em Nature Reviews Drug Discovery}, vol.~18, no.~6, pp.~463--477, 2019.

\bibitem{kitchin_machine_2018}
J.~R. Kitchin, ``Machine learning in catalysis,'' {\em Nature Catalysis}, vol.~1, no.~4, pp.~230--232, 2018.

\bibitem{toyao_machine_2020}
T.~Toyao, Z.~Maeno, S.~Takakusagi, T.~Kamachi, I.~Takigawa, and K.-i. Shimizu, ``Machine {Learning} for {Catalysis} {Informatics}: {Recent} {Applications} and {Prospects},'' {\em ACS Catalysis}, vol.~10, no.~3, pp.~2260--2297, 2020.

\bibitem{morgan_opportunities_2020}
D.~Morgan and R.~Jacobs, ``Opportunities and {Challenges} for {Machine} {Learning} in {Materials} {Science},'' {\em Annual Review of Materials Research}, vol.~50, no.~Volume 50, 2020, pp.~71--103, 2020.

\bibitem{wei_machine_2019}
J.~Wei, X.~Chu, X.-Y. Sun, K.~Xu, H.-X. Deng, J.~Chen, Z.~Wei, and M.~Lei, ``Machine learning in materials science,'' {\em InfoMat}, vol.~1, no.~3, pp.~338--358, 2019.

\bibitem{hanna_machine-learning_2020}
B.~N. Hanna, N.~T. Dinh, R.~W. Youngblood, and I.~A. Bolotnov, ``Machine-learning based error prediction approach for coarse-grid {Computational} {Fluid} {Dynamics} ({CG}-{CFD}),'' {\em Progress in Nuclear Energy}, vol.~118, p.~103140, 2020.

\bibitem{kochkov_machine_2021}
D.~Kochkov, J.~A. Smith, A.~Alieva, Q.~Wang, M.~P. Brenner, and S.~Hoyer, ``Machine learning–accelerated computational fluid dynamics,'' {\em Proceedings of the National Academy of Sciences}, vol.~118, no.~21, p.~e2101784118, 2021.

\bibitem{gastegger_machine_2017}
M.~Gastegger, J.~Behler, and P.~Marquetand, ``Machine learning molecular dynamics for the simulation of infrared spectra,'' {\em Chemical Science}, vol.~8, no.~10, pp.~6924--6935, 2017.

\bibitem{wang_machine_2020}
Y.~Wang, J.~M. Lamim~Ribeiro, and P.~Tiwary, ``Machine learning approaches for analyzing and enhancing molecular dynamics simulations,'' {\em Current Opinion in Structural Biology}, vol.~61, pp.~139--145, 2020.

\bibitem{arunthavanathan_machine_2022}
R.~Arunthavanathan, S.~Ahmed, F.~Khan, and S.~Imtiaz, ``Machine {Learning} for {Process} {Fault} {Detection} and {Diagnosis},'' in {\em Machine {Learning} in {Chemical} {Safety} and {Health}}, pp.~113--137, John Wiley \& Sons, Ltd, 2022.

\bibitem{harkat_machine_2020}
M.-F. Harkat, A.~Kouadri, R.~Fezai, M.~Mansouri, H.~Nounou, and M.~Nounou, ``Machine {Learning}-{Based} {Reduced} {Kernel} {PCA} {Model} for {Nonlinear} {Chemical} {Process} {Monitoring},'' {\em Journal of Control, Automation and Electrical Systems}, vol.~31, no.~5, pp.~1196--1209, 2020.

\bibitem{kumari_development_2021}
P.~Kumari, B.~Bhadriraju, Q.~Wang, and J.~S.-I. Kwon, ``Development of parametric reduced-order model for consequence estimation of rare events,'' {\em Chemical Engineering Research and Design}, vol.~169, pp.~142--152, 2021.

\bibitem{sarkar_application_2019}
S.~Sarkar, S.~Vinay, R.~Raj, J.~Maiti, and P.~Mitra, ``Application of optimized machine learning techniques for prediction of occupational accidents,'' {\em Computers \& Operations Research}, vol.~106, pp.~210--224, 2019.

\bibitem{tamascelli_learning_2022}
N.~Tamascelli, R.~Solini, N.~Paltrinieri, and V.~Cozzani, ``Learning from major accidents: {A} machine learning approach,'' {\em Computers \& Chemical Engineering}, vol.~162, p.~107786, 2022.

\bibitem{quatrini_machine_2020}
E.~Quatrini, F.~Costantino, G.~Di~Gravio, and R.~Patriarca, ``Machine learning for anomaly detection and process phase classification to improve safety and maintenance activities,'' {\em Journal of Manufacturing Systems}, vol.~56, pp.~117--132, 2020.

\bibitem{budach_effects_2022}
L.~Budach, M.~Feuerpfeil, N.~Ihde, A.~Nathansen, N.~Noack, H.~Patzlaff, F.~Naumann, and H.~Harmouch, ``The {Effects} of {Data} {Quality} on {Machine} {Learning} {Performance},'' 2022.
\newblock arXiv:2207.14529 [cs].

\bibitem{jain_overview_2020}
A.~Jain, H.~Patel, L.~Nagalapatti, N.~Gupta, S.~Mehta, S.~Guttula, S.~Mujumdar, S.~Afzal, R.~Sharma~Mittal, and V.~Munigala, ``Overview and {Importance} of {Data} {Quality} for {Machine} {Learning} {Tasks},'' in {\em Proceedings of the 26th {ACM} {SIGKDD} {International} {Conference} on {Knowledge} {Discovery} \& {Data} {Mining}}, {KDD} '20, (New York, NY, USA), pp.~3561--3562, Association for Computing Machinery, 2020.

\bibitem{wu_moleculenet_2018}
Z.~Wu, B.~Ramsundar, E.~N. Feinberg, J.~Gomes, C.~Geniesse, A.~S. Pappu, K.~Leswing, and V.~Pande, ``{MoleculeNet}: a benchmark for molecular machine learning,'' {\em Chemical Science}, vol.~9, no.~2, pp.~513--530, 2018.

\bibitem{stallkamp_man_2012}
J.~Stallkamp, M.~Schlipsing, J.~Salmen, and C.~Igel, ``Man vs. computer: {Benchmarking} machine learning algorithms for traffic sign recognition,'' {\em Neural Networks}, vol.~32, pp.~323--332, 2012.

\bibitem{purushotham_benchmarking_2018}
S.~Purushotham, C.~Meng, Z.~Che, and Y.~Liu, ``Benchmarking deep learning models on large healthcare datasets,'' {\em Journal of Biomedical Informatics}, vol.~83, pp.~112--134, 2018.

\bibitem{he_fedml_2020}
C.~He, S.~Li, J.~So, X.~Zeng, M.~Zhang, H.~Wang, X.~Wang, P.~Vepakomma, A.~Singh, H.~Qiu, X.~Zhu, J.~Wang, L.~Shen, P.~Zhao, Y.~Kang, Y.~Liu, R.~Raskar, Q.~Yang, M.~Annavaram, and S.~Avestimehr, ``{FedML}: {A} {Research} {Library} and {Benchmark} for {Federated} {Machine} {Learning},'' 2020.
\newblock arXiv:2007.13518 [cs, stat].

\bibitem{thiyagalingam_scientific_2022}
J.~Thiyagalingam, M.~Shankar, G.~Fox, and T.~Hey, ``Scientific machine learning benchmarks,'' {\em Nature Reviews Physics}, vol.~4, no.~6, pp.~413--420, 2022.

\bibitem{aleem_benchmarking_2015}
S.~Aleem, L.~F. Capretz, and F.~Ahmed, ``Benchmarking {Machine} {Learning} {Technologies} for {Software} {Defect} {Detection},'' 2015.
\newblock arXiv:1506.07563 [cs].

\bibitem{pfisterer_benchmarking_2021}
F.~Pfisterer, L.~Beggel, X.~Sun, F.~Scheipl, and B.~Bischl, ``Benchmarking time series classification -- {Functional} data vs machine learning approaches,'' 2021.
\newblock arXiv:1911.07511 [cs, stat].

\bibitem{xie_benchmarking_2020}
J.~Xie and Q.~Wang, ``Benchmarking {Machine} {Learning} {Algorithms} on {Blood} {Glucose} {Prediction} for {Type} {I} {Diabetes} in {Comparison} {With} {Classical} {Time}-{Series} {Models},'' {\em IEEE Transactions on Biomedical Engineering}, vol.~67, no.~11, pp.~3101--3124, 2020.

\bibitem{feltes_cumida_2019}
B.~C. Feltes, E.~B. Chandelier, B.~I. Grisci, and M.~Dorn, ``{CuMiDa}: {An} {Extensively} {Curated} {Microarray} {Database} for {Benchmarking} and {Testing} of {Machine} {Learning} {Approaches} in {Cancer} {Research},'' {\em Journal of Computational Biology}, vol.~26, no.~4, pp.~376--386, 2019.

\bibitem{shwartz-ziv_tabular_2022}
R.~Shwartz-Ziv and A.~Armon, ``Tabular data: {Deep} learning is not all you need,'' {\em Information Fusion}, vol.~81, pp.~84--90, 2022.

\bibitem{borisov_deep_2022}
V.~Borisov, T.~Leemann, K.~Seßler, J.~Haug, M.~Pawelczyk, and G.~Kasneci, ``Deep {Neural} {Networks} and {Tabular} {Data}: {A} {Survey},'' {\em IEEE Transactions on Neural Networks and Learning Systems}, pp.~1--21, 2022.

\bibitem{grinsztajn_why_2022}
L.~Grinsztajn, E.~Oyallon, and G.~Varoquaux, ``Why do tree-based models still outperform deep learning on typical tabular data?,'' {\em Advances in Neural Information Processing Systems}, vol.~35, pp.~507--520, 2022.

\bibitem{uddin_confirming_2024}
S.~Uddin and H.~Lu, ``Confirming the statistically significant superiority of tree-based machine learning algorithms over their counterparts for tabular data,'' {\em PLOS ONE}, vol.~19, no.~4, p.~e0301541, 2024.

\bibitem{shyalika_comprehensive_2023}
C.~Shyalika, R.~Wickramarachchi, and A.~Sheth, ``A {Comprehensive} {Survey} on {Rare} {Event} {Prediction},'' 2023.
\newblock arXiv:2309.11356 [cs].

\bibitem{sudarshan_understanding_2021}
V.~Sudarshan, W.~D. Seider, A.~J. Patel, and J.~E. Arbogast, ``Understanding rare safety and reliability events using forward-flux sampling,'' {\em Computers \& Chemical Engineering}, vol.~153, p.~107387, 2021.

\bibitem{sudarshan_multivariate_2023}
V.~Sudarshan, W.~D. Seider, A.~J. Patel, U.~G. Oktem, and J.~E. Arbogast, ``Multivariate alarm systems to recognize rare unpostulated abnormal events,'' {\em AIChE Journal}, vol.~70, no.~2, p.~e18284, 2023.

\bibitem{sudarshan_alarm_2024}
V.~Sudarshan, W.~D. Seider, A.~J. Patel, U.~G. Oktem, and J.~E. Arbogast, ``Alarm rationalization and dynamic risk analyses for rare abnormal events,'' {\em Computers \& Chemical Engineering}, vol.~184, p.~108633, 2024.

\bibitem{sudarshan_path-sampling_2024}
V.~Sudarshan, W.~D. Seider, A.~J. Patel, U.~G. Oktem, and J.~E. Arbogast, ``Path-{Sampling} and {Machine} learning for rare abnormal {Events}: {Application} to polymerization {CSTRs},'' {\em Chemical Engineering Science}, p.~120513, 2024.

\bibitem{filion_crystal_2010}
L.~Filion, M.~Hermes, R.~Ni, and M.~Dijkstra, ``Crystal nucleation of hard spheres using molecular dynamics, umbrella sampling, and forward flux sampling: {A} comparison of simulation techniques,'' {\em The Journal of Chemical Physics}, vol.~133, no.~24, p.~244115, 2010.

\bibitem{jiang_forward_2018}
H.~Jiang, A.~Haji-Akbari, P.~G. Debenedetti, and A.~Z. Panagiotopoulos, ``Forward flux sampling calculation of homogeneous nucleation rates from aqueous {NaCl} solutions,'' {\em The Journal of Chemical Physics}, vol.~148, no.~4, p.~044505, 2018.

\bibitem{allen_simulating_2006}
R.~J. Allen, D.~Frenkel, and P.~R. ten Wolde, ``Simulating rare events in equilibrium or nonequilibrium stochastic systems,'' {\em The Journal of Chemical Physics}, vol.~124, no.~2, p.~024102, 2006.

\bibitem{arjun_homogeneous_2023}
A.~Arjun and P.~G. Bolhuis, ``Homogeneous nucleation of crystalline methane hydrate in molecular dynamics transition paths sampled under realistic conditions,'' {\em The Journal of Chemical Physics}, vol.~158, no.~4, p.~044504, 2023.

\bibitem{bi_probing_2014}
Y.~Bi and T.~Li, ``Probing {Methane} {Hydrate} {Nucleation} through the {Forward} {Flux} {Sampling} {Method},'' {\em The Journal of Physical Chemistry B}, vol.~118, no.~47, pp.~13324--13332, 2014.

\bibitem{moskowitz_understanding_2018}
I.~H. Moskowitz, W.~D. Seider, A.~J. Patel, J.~E. Arbogast, and U.~G. Oktem, ``Understanding rare safety and reliability events using transition path sampling,'' {\em Computers \& Chemical Engineering}, vol.~108, pp.~74--88, 2018.

\bibitem{bolhuis_transition_2002}
P.~G. Bolhuis, D.~Chandler, C.~Dellago, and P.~L. Geissler, ``Transition {Path} {Sampling} : {Throwing} {Ropes} {Over} {Rough} {Mountain} {Passes}, in the {Dark},'' {\em Annual Review of Physical Chemistry}, vol.~53, no.~1, pp.~291--318, 2002.

\bibitem{dellago_transition_1998}
C.~Dellago, P.~G. Bolhuis, F.~S. Csajka, and D.~Chandler, ``Transition path sampling and the calculation of rate constants,'' {\em The Journal of Chemical Physics}, vol.~108, no.~5, pp.~1964--1977, 1998.

\bibitem{prigogine_transition_2002}
C.~Dellago, P.~G. Bolhuis, and P.~L. Geissler, ``Transition {Path} {Sampling},'' in {\em Advances in {Chemical} {Physics}}, vol.~123, pp.~1--78, Wiley, 1~ed., 2002.

\bibitem{peters_obtaining_2006}
B.~Peters and B.~L. Trout, ``Obtaining reaction coordinates by likelihood maximization,'' {\em The Journal of Chemical Physics}, vol.~125, no.~5, p.~054108, 2006.

\bibitem{borrero_reaction_2007}
E.~E. Borrero and F.~A. Escobedo, ``Reaction coordinates and transition pathways of rare events via forward flux sampling,'' {\em The Journal of Chemical Physics}, vol.~127, no.~16, p.~164101, 2007.

\bibitem{bichri_investigating_2024}
H.~Bichri, A.~Chergui, and M.~Hain, ``Investigating the {Impact} of {Train} / {Test} {Split} {Ratio} on the {Performance} of {Pre}-{Trained} {Models} with {Custom} {Datasets},'' {\em International Journal of Advanced Computer Science and Applications}, vol.~15, no.~2, 2024.

\bibitem{kahloot_algorithmic_2021}
K.~M. Kahloot and P.~Ekler, ``Algorithmic {Splitting}: {A} {Method} for {Dataset} {Preparation},'' {\em IEEE Access}, vol.~9, pp.~125229--125237, 2021.

\bibitem{vrigazova_proportion_2021}
B.~Vrigazova, ``The {Proportion} for {Splitting} {Data} into {Training} and {Test} {Set} for the {Bootstrap} in {Classification} {Problems},'' {\em Business Systems Research : International journal of the Society for Advancing Innovation and Research in Economy}, vol.~12, no.~1, pp.~228--242, 2021.

\bibitem{bergstra_hyperopt_2015}
J.~Bergstra, B.~Komer, C.~Eliasmith, D.~Yamins, and D.~D. Cox, ``Hyperopt: a {Python} library for model selection and hyperparameter optimization,'' {\em Computational Science \& Discovery}, vol.~8, no.~1, p.~014008, 2015.

\bibitem{akiba_optuna_2019}
T.~Akiba, S.~Sano, T.~Yanase, T.~Ohta, and M.~Koyama, ``Optuna: {A} {Next}-generation {Hyperparameter} {Optimization} {Framework},'' 2019.
\newblock arXiv:1907.10902 [cs, stat].

\bibitem{liaw_tune_2018}
R.~Liaw, E.~Liang, R.~Nishihara, P.~Moritz, J.~E. Gonzalez, and I.~Stoica, ``Tune: {A} {Research} {Platform} for {Distributed} {Model} {Selection} and {Training},'' 2018.
\newblock arXiv:1807.05118 [cs, stat].

\bibitem{claesen_easy_2014}
M.~Claesen, J.~Simm, D.~Popovic, Y.~Moreau, and B.~De~Moor, ``Easy {Hyperparameter} {Search} {Using} {Optunity},'' 2014.
\newblock arXiv:1412.1114 [cs].

\bibitem{bergstra_algorithms_2011}
J.~Bergstra, R.~Bardenet, Y.~Bengio, and B.~Kégl, ``Algorithms for {Hyper}-{Parameter} {Optimization},'' in {\em Advances in {Neural} {Information} {Processing} {Systems}}, vol.~24, Curran Associates, Inc., 2011.

\bibitem{watanabe_tree-structured_2023}
S.~Watanabe, ``Tree-{Structured} {Parzen} {Estimator}: {Understanding} {Its} {Algorithm} {Components} and {Their} {Roles} for {Better} {Empirical} {Performance},'' 2023.
\newblock arXiv:2304.11127 [cs].

\bibitem{motz_benchmarking_2022}
M.~Motz, J.~Krauß, and R.~H. Schmitt, ``Benchmarking of hyperparameter optimization techniques for machine learning applications in production,'' {\em Advances in Industrial and Manufacturing Engineering}, vol.~5, p.~100099, 2022.

\bibitem{shekhar_comparative_2022}
S.~Shekhar, A.~Bansode, and A.~Salim, ``A {Comparative} study of {Hyper}-{Parameter} {Optimization} {Tools},'' 2022.
\newblock arXiv:2201.06433 [cs.LG].

\bibitem{cortes_support-vector_1995}
C.~Cortes and V.~Vapnik, ``Support-vector networks,'' {\em Machine Learning}, vol.~20, no.~3, pp.~273--297, 1995.

\bibitem{drucker_support_1996}
H.~Drucker, C.~J.~C. Burges, L.~Kaufman, A.~Smola, and V.~Vapnik, ``Support {Vector} {Regression} {Machines},'' in {\em Advances in {Neural} {Information} {Processing} {Systems}}, vol.~9, MIT Press, 1996.

\bibitem{danielsson_euclidean_1980}
P.-E. Danielsson, ``Euclidean distance mapping,'' {\em Computer Graphics and Image Processing}, vol.~14, no.~3, pp.~227--248, 1980.

\bibitem{chen_comparative_2011}
Z.~Li, Q.~Ding, and W.~Zhang, ``A {Comparative} {Study} of {Different} {Distances} for {Similarity} {Estimation},'' in {\em Intelligent {Computing} and {Information} {Science}}, vol.~134, pp.~483--488, Berlin, Heidelberg: Springer Berlin Heidelberg, 2011.
\newblock Series Title: Communications in Computer and Information Science.

\bibitem{suwanda_analysis_2020}
R.~Suwanda, Z.~Syahputra, and E.~M. Zamzami, ``Analysis of {Euclidean} {Distance} and {Manhattan} {Distance} in the {K}-{Means} {Algorithm} for {Variations} {Number} of {Centroid} {K},'' {\em Journal of Physics: Conference Series}, vol.~1566, no.~1, p.~012058, 2020.

\bibitem{breiman_random_2001}
L.~Breiman, ``Random {Forests},'' {\em Machine Learning}, vol.~45, no.~1, pp.~5--32, 2001.

\bibitem{scikit-learn_developers_randomforestregressor_2024}
scikit-learn developers, ``{RandomForestRegressor},'' 2024.

\bibitem{chen_xgboost_2016}
T.~Chen and C.~Guestrin, ``{XGBoost}: {A} {Scalable} {Tree} {Boosting} {System},'' in {\em Proceedings of the 22nd {ACM} {SIGKDD} {International} {Conference} on {Knowledge} {Discovery} and {Data} {Mining}}, pp.~785--794, 2016.
\newblock arXiv:1603.02754 [cs].

\bibitem{cerna_comparison_2020}
S.~Cerna, C.~Guyeux, H.~H. Arcolezi, R.~Couturier, and G.~Royer, ``A {Comparison} of {LSTM} and {XGBoost} for {Predicting} {Firemen} {Interventions},'' in {\em Trends and {Innovations} in {Information} {Systems} and {Technologies}}, Advances in {Intelligent} {Systems} and {Computing}, (Cham), pp.~424--434, Springer International Publishing, 2020.

\bibitem{li_gene_2019}
W.~Li, Y.~Yin, X.~Quan, and H.~Zhang, ``Gene {Expression} {Value} {Prediction} {Based} on {XGBoost} {Algorithm},'' {\em Frontiers in Genetics}, vol.~10, 2019.

\bibitem{ma_xgboost-based_2021}
M.~Ma, G.~Zhao, B.~He, Q.~Li, H.~Dong, S.~Wang, and Z.~Wang, ``{XGBoost}-based method for flash flood risk assessment,'' {\em Journal of Hydrology}, vol.~598, p.~126382, 2021.

\bibitem{ogunleye_xgboost_2020}
A.~Ogunleye and Q.-G. Wang, ``{XGBoost} {Model} for {Chronic} {Kidney} {Disease} {Diagnosis},'' {\em IEEE/ACM Transactions on Computational Biology and Bioinformatics}, vol.~17, no.~6, pp.~2131--2140, 2020.

\bibitem{ashraf_identification_2023}
M.~T. Ashraf, K.~Dey, and S.~Mishra, ``Identification of high-risk roadway segments for wrong-way driving crash using rare event modeling and data augmentation techniques,'' {\em Accident Analysis \& Prevention}, vol.~181, p.~106933, 2023.

\bibitem{wang_classification_2023}
T.~Wang, Y.~Bian, Y.~Zhang, and X.~Hou, ``Classification of earthquakes, explosions and mining-induced earthquakes based on {XGBoost} algorithm,'' {\em Computers \& Geosciences}, vol.~170, p.~105242, 2023.

\bibitem{xgboost_developers_xgboost_2023}
xgboost developers, ``xgboost {Release} 1.7.6,'' 2023.

\bibitem{ke_lightgbm_2017}
G.~Ke, Q.~Meng, T.~Finley, T.~Wang, W.~Chen, W.~Ma, Q.~Ye, and T.-Y. Liu, ``{LightGBM}: {A} {Highly} {Efficient} {Gradient} {Boosting} {Decision} {Tree},'' in {\em Advances in {Neural} {Information} {Processing} {Systems}}, vol.~30, Curran Associates, Inc., 2017.

\bibitem{liang_predicting_2020}
W.~Liang, S.~Luo, G.~Zhao, and H.~Wu, ``Predicting {Hard} {Rock} {Pillar} {Stability} {Using} {GBDT}, {XGBoost}, and {LightGBM} {Algorithms},'' {\em Mathematics}, vol.~8, no.~5, p.~765, 2020.

\bibitem{prokhorenkova_catboost_2018}
L.~Prokhorenkova, G.~Gusev, A.~Vorobev, A.~V. Dorogush, and A.~Gulin, ``{CatBoost}: unbiased boosting with categorical features,'' in {\em Advances in {Neural} {Information} {Processing} {Systems}}, vol.~31, Curran Associates, Inc., 2018.

\bibitem{daoud_comparison_2019}
E.~A. Daoud, ``Comparison between {XGBoost}, {LightGBM} and {CatBoost} {Using} a {Home} {Credit} {Dataset},'' {\em International Journal of Computer and Information Engineering}, vol.~13, no.~1, pp.~6--10, 2019.

\bibitem{so_enhanced_2024}
B.~So, ``Enhanced gradient boosting for zero-inflated insurance claims and comparative analysis of {CatBoost}, {XGBoost}, and {LightGBM},'' {\em Scandinavian Actuarial Journal}, vol.~0, no.~0, pp.~1--23, 2024.

\bibitem{hancock_performance_2020}
J.~Hancock and T.~M. Khoshgoftaar, ``Performance of {CatBoost} and {XGBoost} in {Medicare} {Fraud} {Detection},'' in {\em 2020 19th {IEEE} {International} {Conference} on {Machine} {Learning} and {Applications} ({ICMLA})}, pp.~572--579, 2020.

\bibitem{traore_deep_2018}
B.~B. Traore, B.~Kamsu-Foguem, and F.~Tangara, ``Deep convolution neural network for image recognition,'' {\em Ecological Informatics}, vol.~48, pp.~257--268, 2018.

\bibitem{ma_image_2020}
S.~Ma, X.~Zhang, C.~Jia, Z.~Zhao, S.~Wang, and S.~Wang, ``Image and {Video} {Compression} {With} {Neural} {Networks}: {A} {Review},'' {\em IEEE Transactions on Circuits and Systems for Video Technology}, vol.~30, no.~6, pp.~1683--1698, 2020.

\bibitem{chen_dialogsum_2021}
Y.~Chen, Y.~Liu, L.~Chen, and Y.~Zhang, ``{DialogSum}: {A} {Real}-{Life} {Scenario} {Dialogue} {Summarization} {Dataset},'' 2021.
\newblock arXiv:2105.06762 [cs].

\bibitem{singh_machine_2017}
S.~P. Singh, A.~Kumar, H.~Darbari, L.~Singh, A.~Rastogi, and S.~Jain, ``Machine translation using deep learning: {An} overview,'' in {\em 2017 {International} {Conference} on {Computer}, {Communications} and {Electronics} ({Comptelix})}, pp.~162--167, 2017.

\bibitem{dos_santos_deep_2014}
C.~dos Santos and M.~Gatti, ``Deep {Convolutional} {Neural} {Networks} for {Sentiment} {Analysis} of {Short} {Texts},'' in {\em Proceedings of {COLING} 2014, the 25th {International} {Conference} on {Computational} {Linguistics}: {Technical} {Papers}}, (Dublin, Ireland), pp.~69--78, Dublin City University and Association for Computational Linguistics, 2014.

\bibitem{rosenblatt_perceptron_1958}
F.~Rosenblatt, ``The perceptron: {A} probabilistic model for information storage and organization in the brain,'' {\em Psychological Review}, vol.~65, no.~6, pp.~386--408, 1958.

\bibitem{nair_rectified_2010}
V.~Nair and G.~E. Hinton, ``Rectified linear units improve restricted boltzmann machines,'' in {\em Proceedings of the 27th international conference on machine learning ({ICML}-10)}, pp.~807--814, 2010.

\bibitem{ruder_overview_2017}
S.~Ruder, ``An overview of gradient descent optimization algorithms,'' 2017.
\newblock arXiv:1609.04747 [cs].

\bibitem{kingma_adam_2017}
D.~P. Kingma and J.~Ba, ``Adam: {A} {Method} for {Stochastic} {Optimization},'' 2017.
\newblock arXiv:1412.6980 [cs].

\bibitem{zou_overview_2009}
J.~Zou, Y.~Han, and S.-S. So, ``Overview of {Artificial} {Neural} {Networks},'' in {\em Artificial {Neural} {Networks}: {Methods} and {Applications}}, pp.~14--22, Totowa, NJ: Humana Press, 2009.

\bibitem{hassoun_fundamentals_1995}
M.~H. Hassoun, {\em Fundamentals of {Artificial} {Neural} {Networks}}.
\newblock MIT Press, 1995.

\bibitem{arik_tabnet_2020}
S.~O. Arik and T.~Pfister, ``{TabNet}: {Attentive} {Interpretable} {Tabular} {Learning},'' 2020.
\newblock arXiv:1908.07442 [cs, stat].

\bibitem{yan_rainfall_2021}
J.~Yan, T.~Xu, Y.~Yu, and H.~Xu, ``Rainfall {Forecast} {Model} {Based} on the {TabNet} {Model},'' {\em Water}, vol.~13, no.~9, p.~1272, 2021.

\bibitem{borghini_short_2021}
E.~Borghini and C.~Giannetti, ``Short {Term} {Load} {Forecasting} {Using} {TabNet}: {A} {Comparative} {Study} with {Traditional} {State}-of-the-{Art} {Regression} {Models},'' {\em Engineering Proceedings}, vol.~5, no.~1, p.~6, 2021.

\bibitem{joseph_explainable_2022}
L.~P. Joseph, E.~A. Joseph, and R.~Prasad, ``Explainable diabetes classification using hybrid {Bayesian}-optimized {TabNet} architecture,'' {\em Computers in Biology and Medicine}, vol.~151, p.~106178, 2022.

\bibitem{mcdonnell_deep_2023}
K.~McDonnell, F.~Murphy, B.~Sheehan, L.~Masello, and G.~Castignani, ``Deep learning in insurance: {Accuracy} and model interpretability using {TabNet},'' {\em Expert Systems with Applications}, vol.~217, p.~119543, 2023.

\bibitem{russo_operability_1998}
L.~P. Russo and B.~W. Bequette, ``Operability of chemical reactors: multiplicity behavior of a jacketed styrene polymerization reactor,'' {\em Chemical Engineering Science}, vol.~53, no.~1, pp.~27--45, 1998.

\bibitem{harris_array_2020}
C.~R. Harris, K.~J. Millman, S.~J. van~der Walt, R.~Gommers, P.~Virtanen, D.~Cournapeau, E.~Wieser, J.~Taylor, S.~Berg, N.~J. Smith, R.~Kern, M.~Picus, S.~Hoyer, M.~H. van Kerkwijk, M.~Brett, A.~Haldane, J.~F. del Río, M.~Wiebe, P.~Peterson, P.~Gérard-Marchant, K.~Sheppard, T.~Reddy, W.~Weckesser, H.~Abbasi, C.~Gohlke, and T.~E. Oliphant, ``Array programming with {NumPy},'' {\em Nature}, vol.~585, no.~7825, pp.~357--362, 2020.

\bibitem{mckinney_data_2010}
W.~McKinney {\em et~al.}, ``Data structures for statistical computing in python.,'' in {\em SciPy}, vol.~445, pp.~51--56, 2010.

\bibitem{virtanen_scipy_2020}
P.~Virtanen, R.~Gommers, T.~E. Oliphant, M.~Haberland, T.~Reddy, D.~Cournapeau, E.~Burovski, P.~Peterson, W.~Weckesser, J.~Bright, S.~J. van~der Walt, M.~Brett, J.~Wilson, K.~J. Millman, N.~Mayorov, A.~R.~J. Nelson, E.~Jones, R.~Kern, E.~Larson, C.~J. Carey, I.~Polat, Y.~Feng, E.~W. Moore, J.~VanderPlas, D.~Laxalde, J.~Perktold, R.~Cimrman, I.~Henriksen, E.~A. Quintero, C.~R. Harris, A.~M. Archibald, A.~H. Ribeiro, F.~Pedregosa, and P.~van Mulbregt, ``{SciPy} 1.0: fundamental algorithms for scientific computing in {Python},'' {\em Nature Methods}, vol.~17, no.~3, pp.~261--272, 2020.

\bibitem{hunter_matplotlib_2007}
J.~D. Hunter, ``Matplotlib: {A} {2D} {Graphics} {Environment},'' {\em Computing in Science \& Engineering}, vol.~9, no.~3, pp.~90--95, 2007.

\bibitem{pedregosa_scikit-learn_2011}
F.~Pedregosa, G.~Varoquaux, A.~Gramfort, V.~Michel, B.~Thirion, O.~Grisel, M.~Blondel, P.~Prettenhofer, R.~Weiss, V.~Dubourg, J.~Vanderplas, A.~Passos, D.~Cournapeau, M.~Brucher, M.~Perrot, and E.~Duchesnay, ``Scikit-learn: {Machine} {Learning} in {Python},'' {\em Journal of Machine Learning Research}, vol.~12, no.~85, pp.~2825--2830, 2011.

\bibitem{lam_numba_2015}
S.~K. Lam, A.~Pitrou, and S.~Seibert, ``Numba: a {LLVM}-based {Python} {JIT} compiler,'' in {\em Proceedings of the {Second} {Workshop} on the {LLVM} {Compiler} {Infrastructure} in {HPC}}, {LLVM} '15, (New York, NY, USA), pp.~1--6, Association for Computing Machinery, 2015.

\bibitem{paszke_pytorch_2019}
A.~Paszke, S.~Gross, F.~Massa, A.~Lerer, J.~Bradbury, G.~Chanan, T.~Killeen, Z.~Lin, N.~Gimelshein, L.~Antiga, A.~Desmaison, A.~Köpf, E.~Yang, Z.~DeVito, M.~Raison, A.~Tejani, S.~Chilamkurthy, B.~Steiner, L.~Fang, J.~Bai, and S.~Chintala, ``{PyTorch}: {An} {Imperative} {Style}, {High}-{Performance} {Deep} {Learning} {Library},'' 2019.
\newblock arXiv:1912.01703 [cs, stat].

\bibitem{dreamquark_pytorch_tabnet_2019}
Dreamquark, ``pytorch\_tabnet documentation,'' 2019.

\bibitem{nvidia_nvidia_2022}
NVIDIA, P.~Vingelmann, and F.~Fitzek, ``{NVIDIA} {CUDA} {Toolkit} 11.8.90,'' 2022.

\bibitem{rapids_development_team_rapids_2023}
R.~D. Team, ``{RAPIDS}: {Libraries} for {End} to {End} {GPU} {Data} {Science},'' 2023.

\bibitem{breiman_chapter_2017}
L.~Breiman, J.~Friedman, R.~Olshen, and C.~Stone, ``Chapter 8: {Regression} {Trees},'' in {\em Classification and {Regression} {Trees}}, Chapman and Hall/CRC, 2017.

\end{thebibliography}

\end{document}